\documentclass[12pt]{scaleai-paper}
\usepackage[square,sort&compress,numbers]{natbib}

\newif\ifCAMERA
\CAMERAfalse

\usepackage{mathpazo}

\usepackage[scaled=0.92]{helvet} 
\usepackage{fontawesome5}
\usepackage{subcaption}
\usepackage{hyperref}
\usepackage{graphicx}

\usepackage{hyperref}       
\hypersetup{
    colorlinks=true,
	linkcolor=blue,
	filecolor=magenta,      
	urlcolor=blue,
	citecolor=black,
	pdfinfo={
        Title={Agentic Rubrics as Contextual Verifiers for SWE Agents},
        Subject={rubrics},
        Keywords={},
    }
}

\usepackage{latexsym}
\usepackage[T1]{fontenc}
\usepackage[utf8]{inputenc}
\usepackage{microtype}
\usepackage{inconsolata}
\usepackage{graphicx}

\let\svthefootnote\thefootnote
\newcommand\freefootnote[1]{%
  \let\thefootnote\relax%
  \footnotetext{#1}%
  \let\thefootnote\svthefootnote%
}
\makeatletter
\renewcommand\AB@affilsepx{, \protect\Affilfont}
\makeatother

\title{Agentic Rubrics as Contextual Verifiers for SWE Agents}

\author{Mohit Raghavendra$^{1, *}$}
\author{Anisha Gunjal$^{1, *}$}
\author{Bing Liu$^1$}
\author{Yunzhong He$^1$}

\affil{$^1$Scale AI}

\newcommand{\authoremail}{%
  \vspace{-1.5em}
    \faEnvelope\  \texttt{mohit.raghavendra@scale.com} \quad 
    \faGlobe\  \url{https://scale.com/research/agenticrubrics}
}

\usepackage{booktabs}
\usepackage{xcolor}
\usepackage{pifont}
\usepackage{subcaption} 

\usepackage{listings}
\usepackage{xcolor}
\usepackage{makecell}
\usepackage{minted}

\usepackage{tabularx}
\usepackage{booktabs}
\usepackage{array}

\usepackage[most]{tcolorbox} 
\usepackage{listings}
\usepackage{xcolor}
\tcbuselibrary{listings}
\usepackage[normalem]{ulem}
\usepackage{minitoc}
\usepackage{multirow}

\lstdefinelanguage{Diff}{
  basicstyle=\ttfamily\scriptsize,
  breaklines=true,
  columns=fullflexible,
  morecomment=[l][\color{green!60!black}]{+},
  morecomment=[l][\color{red!70!black}]{-},
}

\tcbuselibrary{listings, breakable, skins, magazine} 

\usepackage{textcomp}
\definecolor{orange-web}{RGB}{255, 165, 0}
\definecolor{sagegreen}{RGB}{120, 150, 120}
\usepackage[textsize=scriptsize]{todonotes}

\NewDocumentCommand{\mohit}{ mO{} }{\textcolor{orange}{\textsuperscript{\textit{Mohit}}\textsf{\textbf{\small[#1]}}}}

\NewDocumentCommand{\notes}{ mO{} }{\textcolor{gray}{\textsuperscript{\textit{notes}}\textsf{\textbf{\small[#1]}}}}

\NewDocumentCommand{\anisha}{ mO{} }{\textcolor{blue}{\textsuperscript{\textit{Anisha}}\textsf{\textbf{\small[#1]}}}}

\newtcblisting{longyaml}[1][]{
    enhanced,
    breakable,            
    colback=gray!5,       
    colframe=black,       
    boxsep=5pt,
    listing only,
    listing options={
        basicstyle=\ttfamily\scriptsize, 
        breaklines=true,
        columns=fullflexible,
        literate={•}{\textbullet}1 {—}{--}1 {–}{-}1, 
        language=Gnuplot, 
    },
    title={#1},           
    #1
}
\newcommand{\cmark}{\textcolor{green!60!black}{\ding{51}}}
\newcommand{\xmark}{\textcolor{red}{\ding{55}}}

\usepackage{booktabs}
\usepackage[table]{xcolor}   
\usepackage{makecell}
\usepackage{pifont}          
\usepackage{siunitx}         

\let\oldcite\cite
\let\cite\citet
\let\citep\oldcite

\begin{document}
\maketitle
\begingroup
\renewcommand{\thefootnote}{\fnsymbol{footnote}}
\footnotetext[1]{Equal contribution.}
\endgroup
\authoremail

\begin{abstract}
Verification is critical for improving agents: it provides the reward signal for Reinforcement Learning and enables inference-time gains through Test-Time Scaling (TTS). Despite its importance, verification in \emph{software engineering (SWE)} agent settings often relies on code execution, which can be difficult to scale due to environment setup overhead. Scalable alternatives such as patch classifiers and heuristic methods exist, but they are less grounded in codebase context and harder to interpret. To this end, we explore \textbf{Agentic Rubrics}: an expert agent interacts with the repository to create a context-grounded rubric checklist, and candidate patches are then scored against it without requiring test execution. On SWE-Bench Verified under parallel TTS evaluation, Agentic Rubrics achieve a score of \textbf{54.2\%} on \texttt{Qwen3-Coder-30B-A3B} and \textbf{40.6\%} on \texttt{Qwen3-32B}, with at least a \textbf{+3.5} percentage-point gain over the strongest baseline in our comparison set. We further analyze rubric behavior, showing that rubric scores are consistent with ground-truth tests while also flagging issues that tests do not capture. Our ablations show that agentic context gathering is essential for producing codebase-specific, unambiguous criteria. Together, these results suggest that Agentic Rubrics provide an efficient, scalable, and granular verification signal for SWE agents.


\end{abstract}

\section{Introduction}

\ifCAMERA
\else
    \begin{figure}[!t]
        \centering
        \includegraphics[width=0.5\linewidth]{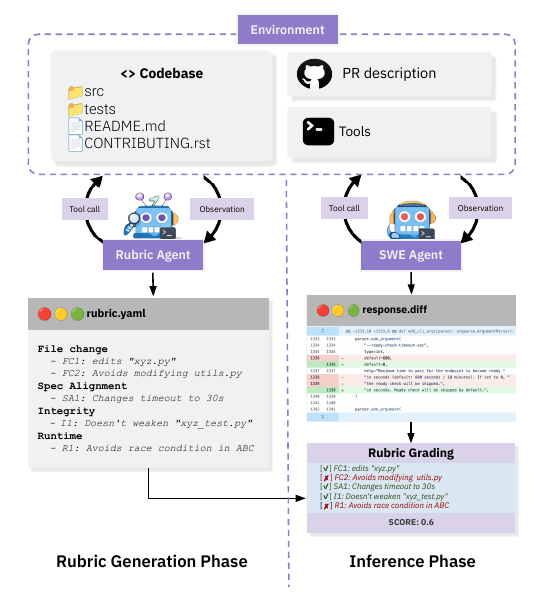}
        \caption{\textbf{Agentic rubric pipeline.} In the rubric-generation phase (left), a rubric agent inspects the codebase and PR description using repository tools, then produces a \texttt{rubric.yaml} organized along four rubric axes (File Change, Spec Alignment, Integrity, Runtime). In the verification phase (right), a SWE agent proposes a patch, which is graded against the rubric to yield an execution-free verifier score.}
        \label{fig:setup}
    \end{figure}
\fi

Large Language Models (LLMs) have rapidly advanced on coding tasks, enabling increasingly capable software engineering (SWE) agents for realistic code editing and bug fixing \citep{yang2024swe, hui2024qwen2, wang2024openhands}. A central bottleneck in training and evaluating such agents is \emph{verification}: determining whether a candidate patch is correct, complete, safe, and aligned with the intended behavior. \emph{Verifier's Law} links the ease of training AI systems on a task to the efficiency and reliability of verifying candidate solutions \citep{wei2025asymmetryverification}. In SWE agent, strong verification plays a dual role. It provides supervision for post-training with verifiable rewards \citep{olmo20252olmo2furious}, and it improves inference through test-time scaling by sampling multiple candidates and selecting the best one using a verifier \citep{brown2024largelanguagemonkeysscaling}.

Current approaches use a range of verifiers, including unit tests (human or LLM-generated), learned patch classifiers, similarity metrics, and LLM judges \citep{luodeepswe, wei2025swerl, jain2025r2e, wei2025trainingsuperintelligentsoftwareagents}. Verification via code execution is environment-aware, but can be costly to scale due to per-instance setup (e.g., sandbox initialization), and may yield sparse or brittle signals, including limited distinguishability and test toxicity \citep{ehrlich2025codemonkeysscalingtesttimecompute, jain2025r2e}. In contrast, execution-free signals are operationally lightweight, but can be less reliable \citep{crupi2025effectivenessllmasajudgecodegeneration}, less interpretable, and prone to shallow cues. As SWE agents expand to more open-ended, goal-driven tasks and long-tail repositories, verifiers must become both scalable and codebase-specific.

To close this gap, we explore \textbf{Agentic Rubrics}. In our setup, illustrated in Figure~\ref{fig:setup}, an expert rubric agent first interacts with a sandboxed repository to synthesize \emph{context-grounded} rubric criteria; after rubric generation, candidate patches are scored without executing code, enabling scalable verification. We build on rubric-based verification \citep{shao2025dr, wu2025rlac}, which decomposes correctness into interpretable criteria that capture partial progress and surface failure modes. For SWE, rubrics written from the problem statement alone are often under-specified because they lack repository-specific context. Our rubric generation is therefore \emph{agentic}: the verifier actively explores the repository to ground criteria in relevant code paths, interfaces, and project conventions, yielding rubric items that are more specific and consistently gradable. We evaluate Agentic Rubrics via best-of-$K$ selection under parallel test-time scaling on SWE-Bench Verified, ablate different design decisions and provide detailed analyses of rubric alignment and utility.

Our contributions are:
(1) We study \textbf{Agentic Rubrics}, a repository-grounded rubric generation paradigm with execution-free scoring for patch selection and post-training.
(2) We show that Agentic Rubrics consistently outperform strong test-based and execution-free verifier baselines under parallel test-time scaling on SWE-Bench Verified. 
(3) We analyze why Agentic Rubrics work, demonstrating alignment with ground-truth tests and showing that rubrics surface diagnostic concerns (e.g., unnecessary edits or missing edge-case handling) even when tests pass.
(4) We demonstrate that agentic rubric generation can be distilled into smaller open-weight models, enabling scalable deployment.

\section{Preliminaries}

\subsection{Verification for SWE Agents} We consider a \emph{verifier} as a procedure that assigns a score to a candidate patch for a given issue, with the goal of selecting or training toward higher-quality solutions. Prior work in SWE Agent settings commonly uses two broad classes of verification signals.
\textbf{Execution-based} methods verify patches by executing code, most often by running unit tests (human-authored ground-truth or LLM-generated) \citep{ehrlich2025codemonkeysscalingtesttimecompute}. \textbf{Execution-free} methods assess patch quality without running the repository, by reranking candidates using learned patch classifiers/verifiers, similarity metrics, or LLM judges \citep{wei2025swerl}.
These approaches occupy different points in the trade-off space between repository grounding, operational cost, and reliability \citep{jain2025r2e}. Execution-based verification is environment-aware but can require per-instance setup (e.g., sandbox initialization) and may yield sparse or brittle signals (e.g., limited distinguishability or test toxicity). Execution-free verification is operationally lightweight, but can be less reliable \citep{crupi2025effectivenessllmasajudgecodegeneration}, less interpretable, and sometimes sensitive to surface-level cues (e.g., stylistic patterns, non-semantic artifacts) rather than functional correctness.

\subsection{Rubric-based Verification}
A \emph{rubric} verifies a candidate patch by decomposing correctness into a small set of explicit criteria \cite{arora2025healthbench}. Concretely, a rubric consists of criteria texts (optionally grouped by axes) with per-criterion weights, and a scoring rule that aggregates criterion-level judgments into a single verifier score. Given a problem and a candidate patch, a judge assigns each criterion a score (e.g., binary or graded) and aggregates them to obtain an overall patch score used for selection or learning. 

For SWE tasks, a key practical consideration is \emph{grounding}. Criteria written solely from the problem statement can omit repository-specific interfaces, constraints, and conventions, which makes judgments less precise and less consistent across patches. This motivates verifiers whose criteria are grounded in the right task-relevant repository context, while still allowing lightweight scoring once criteria are generated.
\section{Experimental Design}

\subsection{Agentic Rubrics}

\paragraph{Rubric Generation.}  We implement a rubric-generation agent on top of the SWE Agent scaffold, which provides tools for repository navigation, file inspection/editing, and shell command execution \citep{yang2024swe, wang2024openhands}. We modify the scaffold's \texttt{SYSTEM PROMPT}, instructing the agent to explore the repository, gather task-relevant context, and produce a patch that adds a structured rubric file, \texttt{rubrics.yaml} (prompt in Appendix~\ref{appendix:rubric_gen_config}). This workflow mirrors how developers validate fixes when comprehensive tests are unavailable: inspecting surrounding code and contracts, tracing call sites, and reasoning about edge cases.

Each rubric item is a tuple $(t_i, w_i)$ consisting of a short natural-language criterion $t_i$ and an importance weight $w_i \in \{1,2,3\}$ (nice-to-have / important / must-have), and is assigned to one of the following \textit{axes}:\\
(i) \textbf{File Change} (4--8 items): edits are minimal, local, and sufficient for the fix;\\
(ii) \textbf{Spec Alignment} (3--6): the patch satisfies the requirements in the issue description;\\
(iii) \textbf{Integrity} (3--6): ``no-cheating'' and hygiene constraints (e.g., no test weakening, broad refactors, mass renames, or dependency churn);\\
(iv) \textbf{Runtime} (3--6): the changes imply the intended runtime behavior and avoid obvious execution-time issues.\\
We parse the submitted \texttt{rubrics.yaml}; if parsing fails, we consider the generation attempt invalid.

\paragraph{Rubric Grading.}
Given a problem and a candidate patch, an LLM judge assigns each rubric item a binary score $s_i \in \{0,1\}$ with importance weight $w_i \in \{1,2,3\}$.
We aggregate scores as $S=\frac{\sum_i w_i s_i}{\sum_i w_i}$, yielding a verifier score $S \in [0,1]$ used for reranking candidates.

\subsection{Test-time Scaling with Agentic Rubrics}
\label{sec:tts}

\begin{figure*}[t]
\centering
\makebox[\textwidth]{%
\begin{minipage}[t]{0.56\textwidth}
  \vspace{0pt}
  \centering
  \scriptsize
  \setlength{\tabcolsep}{3.5pt}
  \renewcommand{\arraystretch}{1.08}

  \begin{tabular}{l | c | c | c c }
  \toprule
  \textbf{Method} &
  \textbf{\makecell{Execution\\Free}} &
  \textbf{\makecell{Expert\\Artifact}} &
  \textbf{\makecell{Qwen3\\32B}} &
  \textbf{\makecell{Qwen3\\Coder}} \\
  \midrule
  Oracle Pass@$16$     & --     & -- & 51.4 & 65.6 \\
  Random Pass@$16$               & --             & -- & 22.6 & 39.6 \\
  \addlinespace[2pt]
  \rowcolor[HTML]{D3DEE1}\multicolumn{5}{l}{\textbf{Non-Agentic Verifiers}} \\
  Self-Consistency   & \cmark             & -- & 33.2 & 47.6 \\
  Patch Classifier     & \cmark             & -- & \underline{37.1} & \underline{50.2} \\
  \addlinespace[2pt]
  \rowcolor[HTML]{DCE5D2}\multicolumn{5}{l}{\textbf{Agentic Verifiers}} \\
  Agentic Tests        & \xmark      & Tests & 33.6 & 49.0 \\  
  Agentic Patch Similarity & \cmark & Patch & 35.0 & 49.6 \\
  \textbf{Agentic Rubrics} (ours) & \cmark & Rubric & \bfseries 40.6 & \bfseries 54.2 \\
  \bottomrule
  \end{tabular}
\end{minipage}\hfill%
\begin{minipage}[t]{0.45\textwidth}
  \vspace{0pt}
  \centering
  \includegraphics[width=\linewidth]{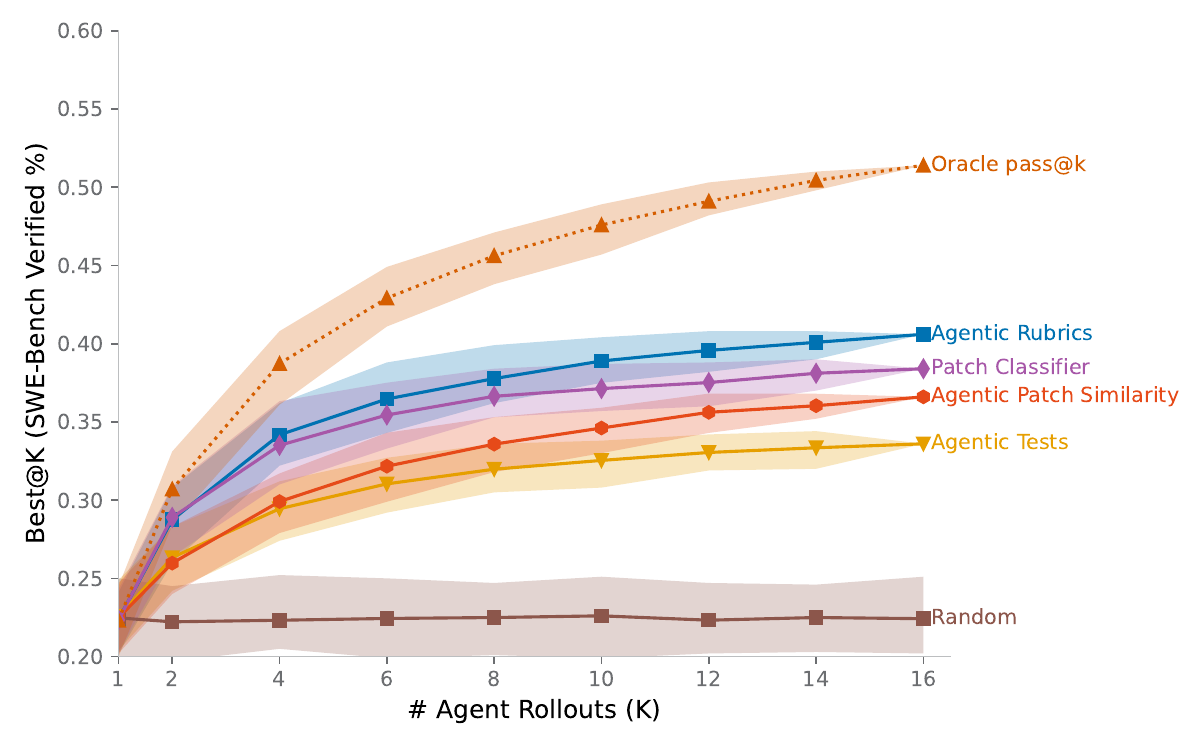}
\end{minipage}%
}
\caption{(Left) Best@16 resolution (\%) with $K=16$ rollouts for \texttt{Qwen3-32B} and \texttt{Qwen3-Coder-30B-A3B}. Verifier signals are generated with \texttt{Claude Sonnet-4.5}; LLM judging uses \texttt{GPT-5 (low reasoning}). (Right) Best@K scaling curves for \texttt{Qwen3-32B} rollouts under different verifiers, with numbers averaged over 100 trials.} 

\label{fig:main_results}
\end{figure*}

\paragraph{Setup}
\label{sec:tts_setup}

Given a SWE problem statement description $D$, a SWE agent produces a rollout trajectory $T^{(j)}$ and candidate patch $P^{(j)}$ for $j = 1, \ldots, K=16$ independent rollouts. The verifier's goal is to assign a score $S^{(j)} \in [0, 1]$ to each candidate. These are then reranked to select the best candidate patch.

\paragraph{Candidate patch generation}
We use the SWE-Agent scaffold by \citet{yang2024swe} as the agent harness for the coding model to interact with the repository in a sandboxed environment to generate patch. For each of the 500 SWE-Bench Verified problems \citep{openai2024swebenchverified}, we sample $16$ independent rollouts and extract candidate patches, from a fixed generator model. We run experiments with two generators: \texttt{Qwen3-32B} (Instruct version, max 30 turns) and \texttt{Qwen3-Coder-30B-A3B} (max 50 turns), both at temperature $1.0$ \citep{yang2025qwen3technicalreport}.

\paragraph{Evaluation protocol (\textsc{Best@}K).}
A problem is considered \emph{resolved} if the candidate patch passes the ground-truth Fail-To-Pass and Pass-to-Pass tests.
We score all $K$ candidates and select the highest-scoring patch for \textsc{Best@K} resolution calculation. In cases where $K<16$, we repeat sample $100$ trials to make this robust. As reference points, we report \textsc{Oracle Pass@K} (selecting using ground-truth tests; an upper bound) and \textsc{Random@K} (uniform selection).

\subsection{Baselines}
\label{sec:tts_baselines}

We group baseline verifiers into two categories based on whether they rely on an \emph{externally generated verification artifact} (e.g., tests, a reference patch, or a rubric) produced via repository interaction. \textbf{Non-agentic verifiers} score candidate patches directly from the problem statement and patch, without generating any additional artifact or inspecting the repository. \textbf{Agentic verifiers} first interact with the repository via an agentic scaffold to produce an artifact that is then used to score and re-rank candidates. Prompts used for all the baselines are provided in Appendix~\ref{sec:agent-prompts}.

\smallskip
\noindent \colorbox[HTML]{D3DEE1}{\textbf{Non-agentic Verifiers (no artifact).}}

(i) \textit{Self-Consistency}~\citep{wang2023selfconsistency, wei2025swerl, singhi2025solve}:
select the candidate patch whose diff has the highest average similarity to the remaining $K{-}1$ candidates.\footnote{We compute similarity using \href{https://docs.python.org/3/library/difflib.html}{\texttt{difflib}}'s \texttt{SequenceMatcher.ratio()} between unified-diff strings.}

(ii) \textit{Patch Classifier}~\citep{pan2024swegym, jain2025r2e}:
an LLM judge predicts patch correctness and outputs a continuous score in $[0,1]$.

\smallskip
\noindent\colorbox[HTML]{DCE5D2}{ \textbf{Agentic Verifiers (artifact-based).}}

(i) \textit{Agentic Tests}~\citep{jain2025r2e}:
an expert agent generates a problem-specific \texttt{test\_issue.py} with repository interaction; candidates are scored by executing these testcases.

(ii) \textit{Agentic Patch Similarity}:
an expert agent generates a context-grounded \emph{proxy reference patch}; candidates are reranked by similarity to this patch (scored by an LLM judge on a 1--5 scale).

(iii) \textbf{\textit{Agentic Rubrics (our method)}}:
an expert agent gathers repository context and synthesizes a structured \texttt{rubrics.yaml}; candidates are graded against rubric criteria to obtain a final verifier score.

\paragraph{Implementation details.} For methods that require a verification artifact (tests, proxy patch, or rubrics), we use \texttt{Claude Sonnet-4.5} as the expert agent (30-turn budget) to generate the artifact via repository interaction; whenever scoring requires an LLM judge (patch classification, similarity scoring, rubric grading), we use \texttt{GPT-5 (low reasoning)}. All verifiers run in the SWE-Bench Verified sandbox with the repository reset to a pre-PR snapshot: agents may inspect the codebase and existing tests, but cannot access or execute the hidden ground-truth evaluation tests, reference patches, or git history. If an artifact wasn't correctly produced (e.g., missing \texttt{test\_issue.py} or invalid \texttt{rubrics.yaml}), we assign a score of 0. Prior work shows that verifier decisions can be unduly influenced by agent's thinking  trace \citep{jain2025r2e}. So to keep evaluation uniform and reduce verifier hacking, scoring for all methods use only on the problem statement, verifier artifact and the final submitted patch, not the full rollout trajectory or tool traces. 

\section{Results}
\label{sec:results}

\subsection{Test Time scaling with Agentic Rubrics}
\label{subsec:tts_results}

\paragraph{Agentic Rubrics improve \textsc{Best@}K selection.}

Figure~\ref{fig:main_results} reports \textsc{Best@16} for rollouts from two generator models, \texttt{Qwen3-32B} and \texttt{Qwen3-Coder}, grouping verifiers into \emph{non-agentic} methods that score patches directly (no artifact) and \emph{agentic} methods that first generate a verification artifact (tests, a proxy patch, or a rubric) via repository interaction. The right panel shows how \textsc{Best@}K scales with \(K\) for \texttt{Qwen3-32B}.

At K{=}16, \textit{Agentic Rubrics} is the top-performing verifier in both settings. For \texttt{Qwen3-32B}, Agentic Rubrics achieves 40.6\% \textsc{Best@16}, improving over the best performing non-agentic baseline (Patch Classifier, 37.1\%) by +3.5 points and over the strongest artifact-based alternative (Agentic Patch Similarity, 35.0\%) by +4.6 points. For \texttt{Qwen3-Coder-30B-A3B}, Agentic Rubrics attains 54.2\%, improving over the best non-agentic baseline (50.2\%) by +4.0 points and over the best agentic baseline (49.6\%) by +4.6 points. The scaling curve further shows that rubric-based scoring maintains an advantage as K increases, indicating that the gain is not confined to a single operating point.

\paragraph{Rubrics provide most effective agentic signal.}
While other agentic baselines also inject repository context, they rely on more brittle intermediate steps. \textit{Agentic Tests} must generate \emph{runnable} tests in the sandbox (including setup/compilation) and those tests must cleanly discriminate between candidates. \textit{Agentic Patch Similarity} scores “closeness” to a proxy reference patch, which can under-rank semantically correct but stylistically different fixes. Rubric artifacts instead use repository interaction to state \emph{what should hold} (file change, spec alignment, integrity, runtime) and then score candidates execution-free against these criteria, yielding a more robust and interpretable scoring signal.

We provide example problems, rubrics, and their grading of responses in Appendix \ref{app:rubric_qualitative_examples}.

\subsection{Analysis of Agentic Rubrics}
\label{sec:rubric_analysis}
We further examine rubric-based verification beyond aggregate \textsc{Best@}K. We first study \emph{score alignment}: whether agentic rubric scores agree with ground-truth tests and patches in \S\ref{subsec:rubric_test_alignment}. We then audit \emph{utility} by categorizing judgments into high- vs.\ low-utility modes, especially when rubrics are stricter than tests, to understand when they add signal beyond the available tests in \S\ref{subsec:rubric_utility}.

\ifCAMERA
\else
    \begin{figure}[t]
        \centering
        \includegraphics[width=0.6\linewidth]{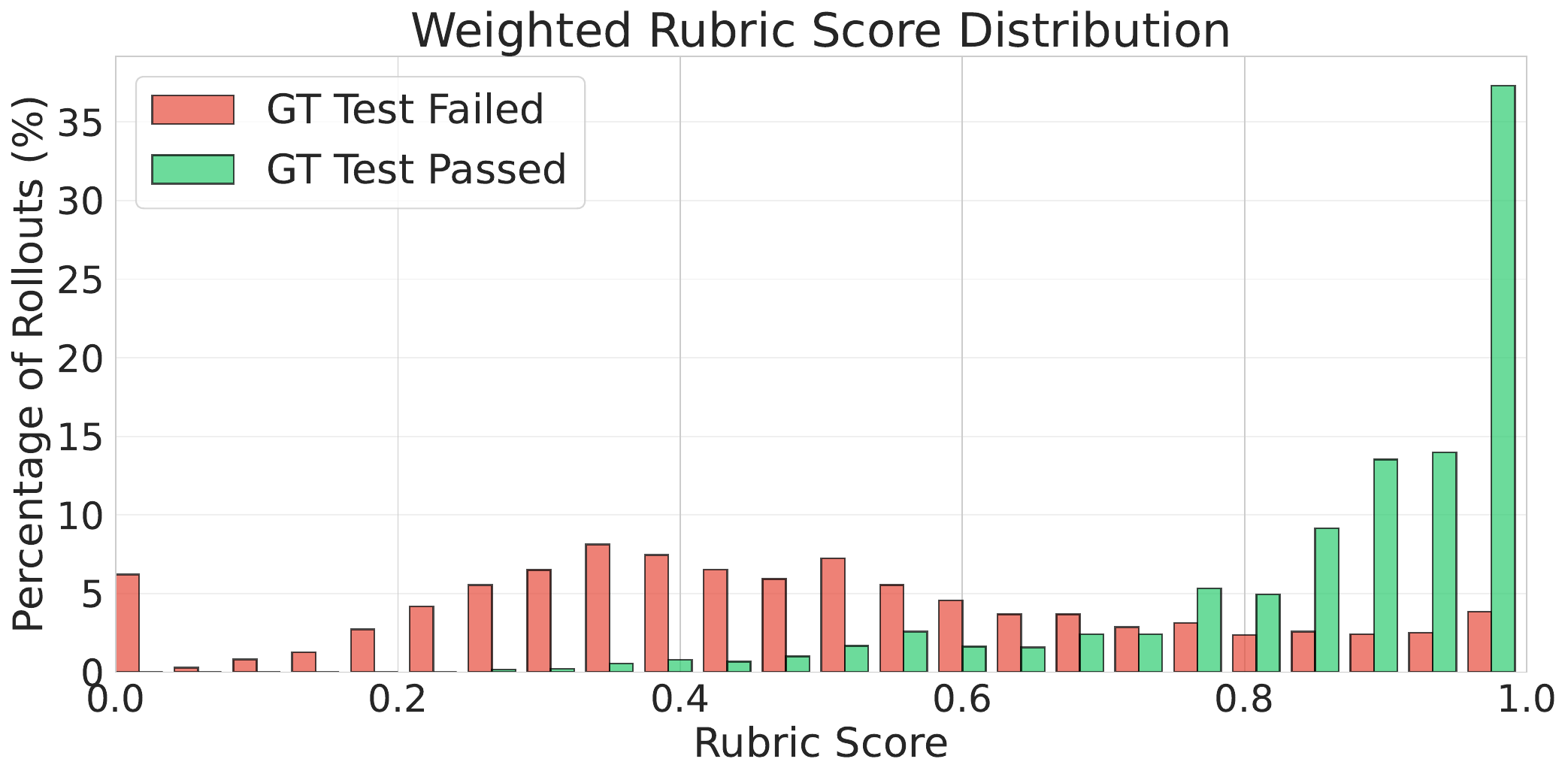}
        \caption{Distribution of Weighted Rubric score for \texttt{Qwen3-32B} rollouts on \texttt{Sonnet-4.5} generated agentic rubrics, for both correct (Ground Truth Tests Pass - Green) and incorrect (Ground-Truth tests Fail - Red). Rubric scores are well aligned with the GT Test correctness signal, awarding lower score for incorrect patches and higher score for correct ones, while providing a denser score distribution.}
        \label{fig:rubric_score_alignment}
    \end{figure}
\fi 

\subsubsection{Rubric Score Alignment Analysis}
\label{subsec:rubric_test_alignment}

\paragraph{Rubric scores separate passing vs.\ failing patches.}
Figure~\ref{fig:rubric_score_alignment} plots \texttt{Sonnet-4.5}'s weighted rubric score distribution on SWE-Bench Verified rollouts by \texttt{Qwen3-32B}, split by whether the candidate patch passes the GT test suite. Rubric scores for GT-Pass rollouts concentrate near high rubric scores (typically 0.85--1.0), while GT-Fail rollouts receive much lower scores on average and spread across a wide range (often around 0.4--0.5). This spread suggests that rubrics can distinguish partial progress from fully-correct solutions, rather than providing only a binary signal like test-suite pass/fail.
Quantitatively, Rubric scores have an ROC-AUC score of \textbf{0.886} and PR-AUC of \textbf{0.722} against GT test Pass/Fail prediction. High PR-AUC suggests that rubrics prioritize true GT-passing patches in the high-precision regime, consistent with providing a more informative graded signal.

Figure~\ref{fig:rubric_category_scores} dissects this across \textit{File Change} (scope), \textit{Spec Alignment}, \textit{Integrity}, and \textit{Runtime}. GT-Failing patches tend to score lower because they make unnecessary edits (\textit{File Change}), miss requirements (\textit{Spec Alignment}), or have runtime issues (\textit{Runtime}), while often remaining strong on \textit{Integrity}. For GT-passing patches, \textit{Spec Alignment} and \textit{Integrity} are near-saturated, but we still see penalties from over-scoped edits and occasional runtime-check issues.

\ifCAMERA
\else
    \begin{figure}[!h]
        \centering
        \includegraphics[width=0.6\linewidth]{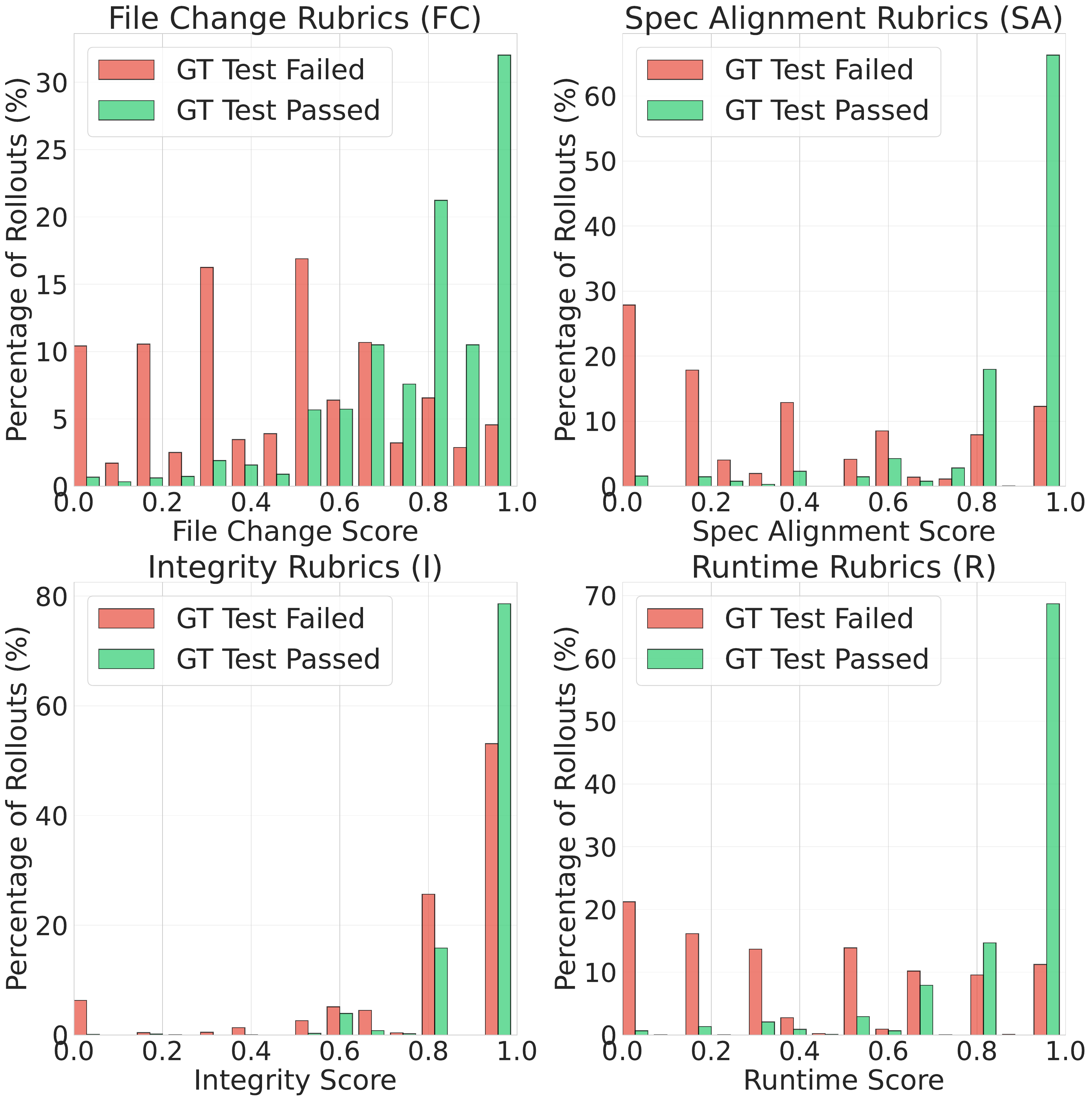}
        \caption{Category-wise distribution of \texttt{Sonnet-4.5} rubric scores on \texttt{Qwen3-32B} rollouts. Incorrect patches (GT Test Failed, in red) score lower on File Change (Edit scope) and Spec Alignment (Satisfying prompt requirements) and Runtime issues, but still good preserving codebase integrity and avoid cheating. Patches that pass ground-truth tests (GT Test Passed, in green) have a very high spec-alignment and integrity score but still suffer from edit scope and in some cases, have issues in runtime checks.}
        \label{fig:rubric_category_scores}
    \end{figure}
\fi

\paragraph{Ground-Truth Patch Agreement} 
We also score human-written \emph{Ground-Truth patches} from the original pull requests, which all pass the SWE-Bench Verified Ground-Truth tests in Appendix \ref{app:rubric_scores_reference}. Table \ref{tab:rubric_scores_reference} and Figure~\ref{fig:reference_patch_rubric_distribution} shows that GT patches receive consistently high rubric scores (mean $>0.8$) across axes from frontier models, suggesting that expert-generated rubrics are broadly compatible with high-quality human fixes. One recurring exception is the \textit{File Change} axis, where rubrics can be more prescriptive about exact edit location/scope than the GT implementation.

\ifCAMERA
\else
\begin{figure*}[t]
    \centering
    \begin{subfigure}[t]{0.49\linewidth}
        \centering
        \includegraphics[width=\linewidth]{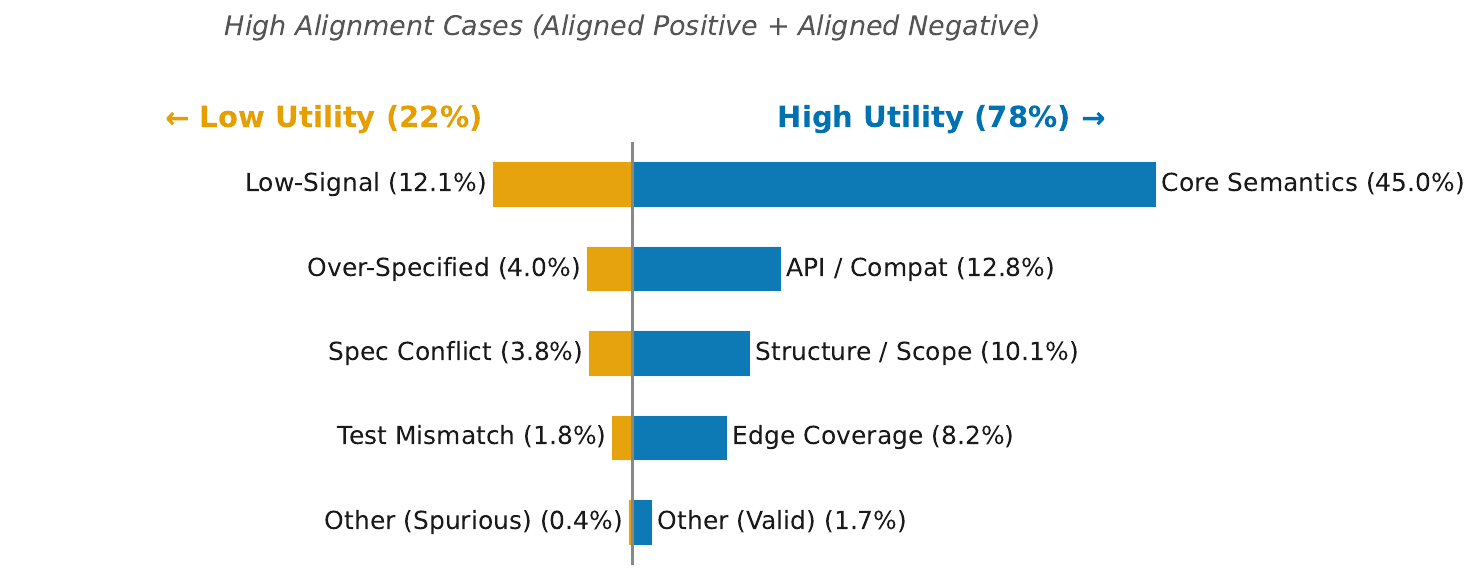}
        \caption{High-alignment cases (ground-truth test case reward = 1, rubric reward $\ge$ 0.7).}
        \label{fig:high_alignment_analysis}
    \end{subfigure}
    \hfill
    \begin{subfigure}[t]{0.49\linewidth}
        \centering
        \includegraphics[width=\linewidth]{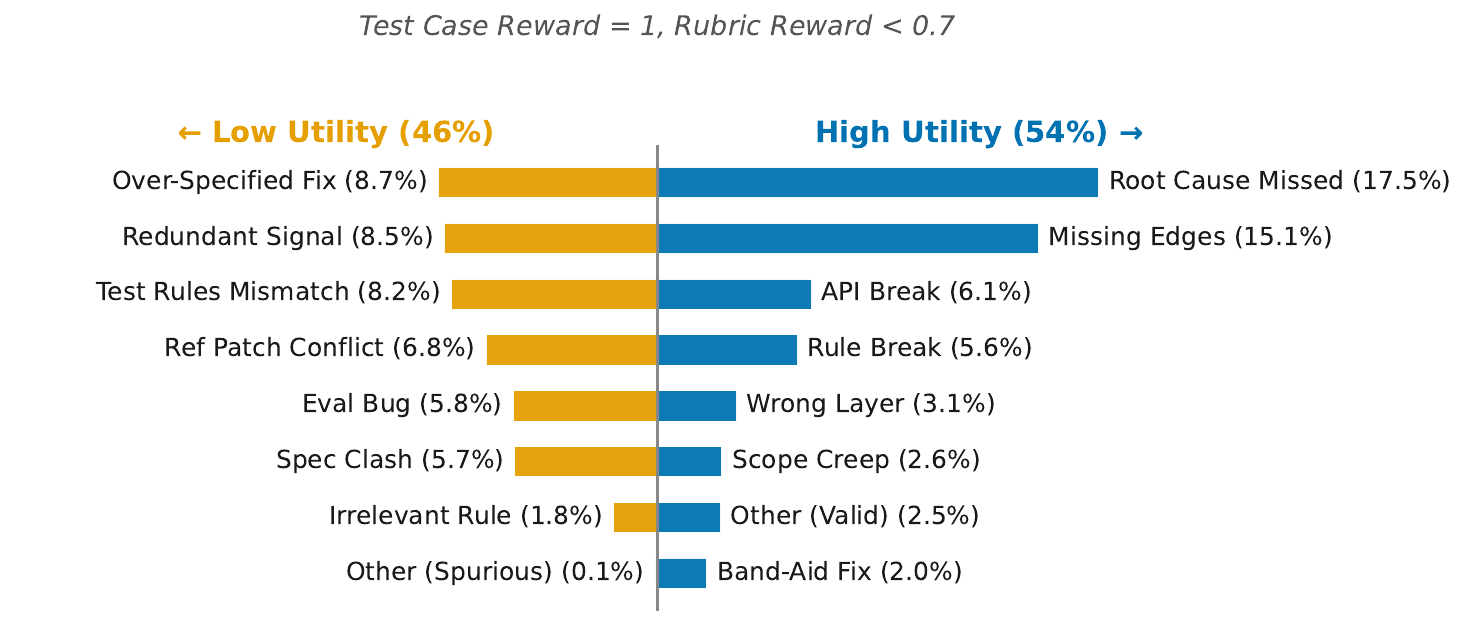}
        \caption{Low-alignment cases (ground-truth test case reward = 1, rubric reward $< 0.7$).}
        \label{fig:low_alignment_analysis}
    \end{subfigure}

    \caption{Qualitative breakdown of agentic rubric utility relative to SWE-Bench Verified ground-truth tests. (a) In high-alignment cases, 78\% of rubrics are high-utility (core semantics, API/compatibility, structure, edge coverage), with 22\% low-utility (low-signal, over-specified, spec- or test-mismatched). (b) When tests pass but rubric scores are low, 54\% of rubric failures are high-utility—often flagging missed root causes or missing edge-case coverage—while 46\% reflect low-utility modes (over-specification, redundancy, rule mismatches, spec conflicts).}
    \label{fig:rubric_utility}
\end{figure*}

\fi

\paragraph{Takeaway}
\emph{Rubric signals are highly correlated with human written Ground Truth tests and patches. Rubric scores also provide a denser signal than test pass/fail by assigning intermediate credit to partially-correct patches and provides detailed technical feedback across different axes.}

\subsubsection{Rubric Utility Analysis}
\label{subsec:rubric_utility}

Since rubrics are generated synthetically without a canonical correctness check, we further scrutinize them for true utility versus spurious signals. We study \emph{when} rubric judgments are useful by labeling them as \emph{high-utility} (spec-consistent, semantically meaningful checks such as core semantics, API/compatibility, structure/scope, edge coverage) or \emph{low-utility} (redundant, over-prescriptive, or misaligned). For a subset of 100 SWE-Bench Verified instances, we prompt \texttt{GPT-5 (medium reasoning)} to assign each case a High-/Low-Utility tag and a sub-category from Table~\ref{tab:rubric_taxonomy}, using the problem statement and Ground-Truth tests and patches as the reference spec (refer Appendix~\ref{sec:utility-prompts}). 

\paragraph{When rubrics and tests agree.}
Figure~\ref{fig:rubric_utility}(a) aggregates cases where rubric and test outcomes agree (both accept or both reject, using a rubric acceptance threshold of $0.7$). In this regime, \textbf{78\%} of rubric judgments are high-utility, primarily reflecting \emph{Core Semantics}, \emph{API/Compatibility}, \emph{Structure/Scope}, and \emph{Edge Coverage}. The remaining \textbf{22\%} are low-utility (e.g., low-signal, over-specified, test-mismatched rubrics, etc.)

\paragraph{When rubrics are stricter than tests.}
Figure~\ref{fig:rubric_utility}(b) considers cases where tests accept a patch but the rubric score is below $0.7$. Even here, \textbf{54\%} of rubric failures are high-utility, most often flagging \emph{Root Cause Missed} and \emph{Missing Edges} - issues that may not be covered by the available GT tests. The remaining \textbf{46\%} low-utility cases highlight failure modes like over-specified fixes, redundant signals, and rubric--test mismatches, which can be mitigated using human-in-the-loop rubric refinement in future work.

\paragraph{Takeaway}
\textit{Across both regimes, most rubric judgments are \emph{substantive}: when rubrics agree with GT-tests, they do so mainly for core semantic and interface reasons (78\% high-utility), \& when rubrics disagree by rejecting GT-test passing patches, over half of these rejections (54\%) flag plausible under-tested issues (root cause missed or missing edges).}

\section{Ablations}

In this section, we ablate key components of the Agentic Rubrics pipeline to isolate the impact of (i) the rubric-agent model, (ii) repository context gathering during rubric construction, and (iii) the judge model used for rubric grading.

\subsection{Rubric-Agent Model Choice}
\label{subsec:model_performance}

\ifCAMERA
\else
    \begin{figure*}[t]
    \centering

    \begin{subfigure}[t]{0.49\linewidth}
        \centering
        \includegraphics[width=\linewidth]{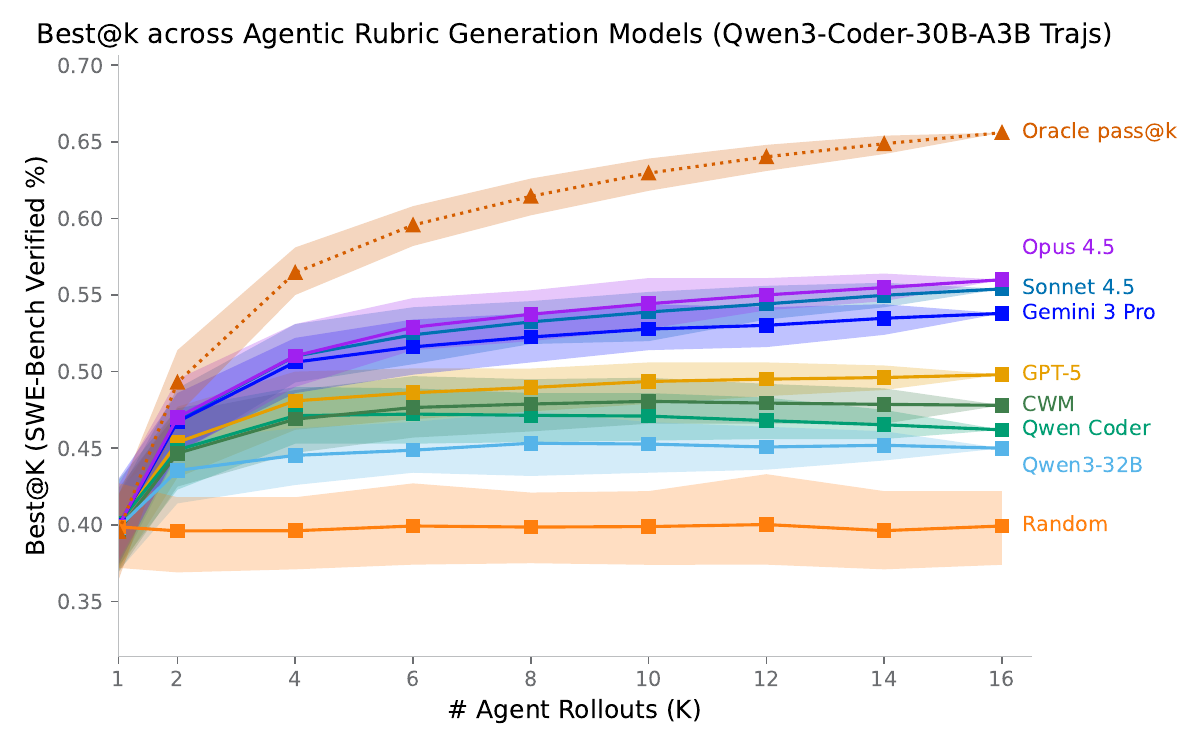}
        \caption{Comparing models as rubric creation agents}
        \label{fig:rubric_agent_models}
    \end{subfigure}
    \hfill
    \begin{subfigure}[t]{0.49\linewidth}
        \centering
        \includegraphics[width=\linewidth]{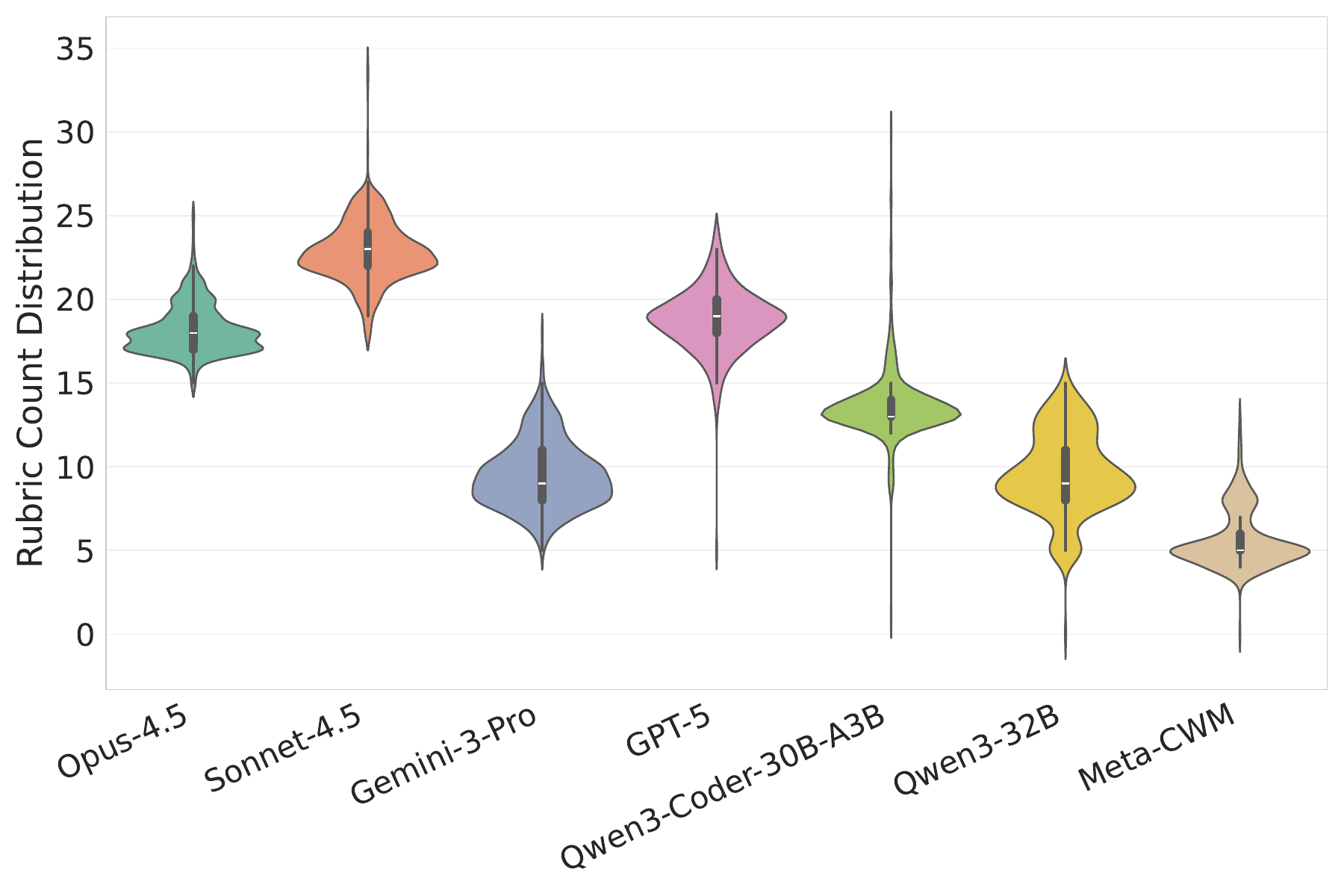}
        \caption{Rubric structure}
        \label{fig:rubric_structure_dist}
    \end{subfigure}

    \caption{(a) Test-time scaling using rubrics generated by various frontier and open models on rollouts from Qwen-Coder-30B-A3B. (b) Distribution of rubric counts per instance across rubric-generation models.}
    \label{fig:model_comparision_rubric_agent}
\end{figure*}

\fi

We investigate the performance of various frontier and Open-weight models in generating rubrics, by studying their test-time performance on rollouts by a fixed policy model (\texttt{Qwen3-Coder-30B-A3B}) and a fixed judge model (\texttt{GPT-5 low reasoning}). The models use their default reasoning effort when applicable. 

Figure \ref{fig:model_comparision_rubric_agent}(a) shows the \textsc{BEST@16} resolution rate when using each model's rubrics on SWE-Bench Verified. We see that the capability of the rubric generation model directly impacts their TTS performance. Frontier coding models (\texttt{Claude Opus-4.5}, \texttt{Claude Sonnet-4.5} and \texttt{Gemini-3-Pro}) achieve the highest \textsc{BEST@16} resolution rates of 54\%. Open-weight coding models like \texttt{Qwen3-Coder-30B-A3B} and \texttt{Code World Model} are not as effective (\~45\%), and finally, non-coding agentic model like \texttt{Qwen3-32B} rubrics yield around 43\%. 
Figure \ref{fig:model_comparision_rubric_agent}(b) provides some insight into this performance gap: more capable models generate sometimes substantially more rubrics per instance. For instance, \texttt{Sonnet-4.5} averages over 20 rubrics per instance, twice that of \texttt{Qwen3-32B} and \texttt{CWM}, although exceptions like \texttt{Gemini-3-Pro} exist. Increased granularity can enable finer-grained differentiation between candidate solutions, explaining the correlation between model capability and selection performance. In addition, rubrics from expert frontier models rubrics are better aligned with ground-truth reference patches, and we analyze this in Appendix \ref{app:rubric_scores_reference}. We also report the model's success rate in using the agentic scaffold and producing parseable rubrics in Appendix \ref{app:agent_scaffold_usage}.

\paragraph{Takeaway} \textit{Rubric-agent capability matters: stronger frontier models generate more granular rubrics with higher human alignment, which translates into higher \textsc{Best@16} performance.}

\subsubsection{Training Open-Weight Rubric Agents}
Motivated by this, we study if such rubric generation capability can be distilled from expert frontier models like \texttt{Sonnet-4.5} into smaller open-models like \texttt{Qwen3-32B}.  We fine-tune \texttt{Qwen3-32B} to \emph{generate agentic rubrics} using the agentic scaffold, and compare against fine-tuning the same model as a \emph{patch classifier} from the same expert model - a common execution-free verifier used in prior works \citep{jain2025r2e, pan2024swegym}.

\ifCAMERA
\else
    \begin{figure}[!h]
        \centering
        \includegraphics[width=0.6\linewidth]{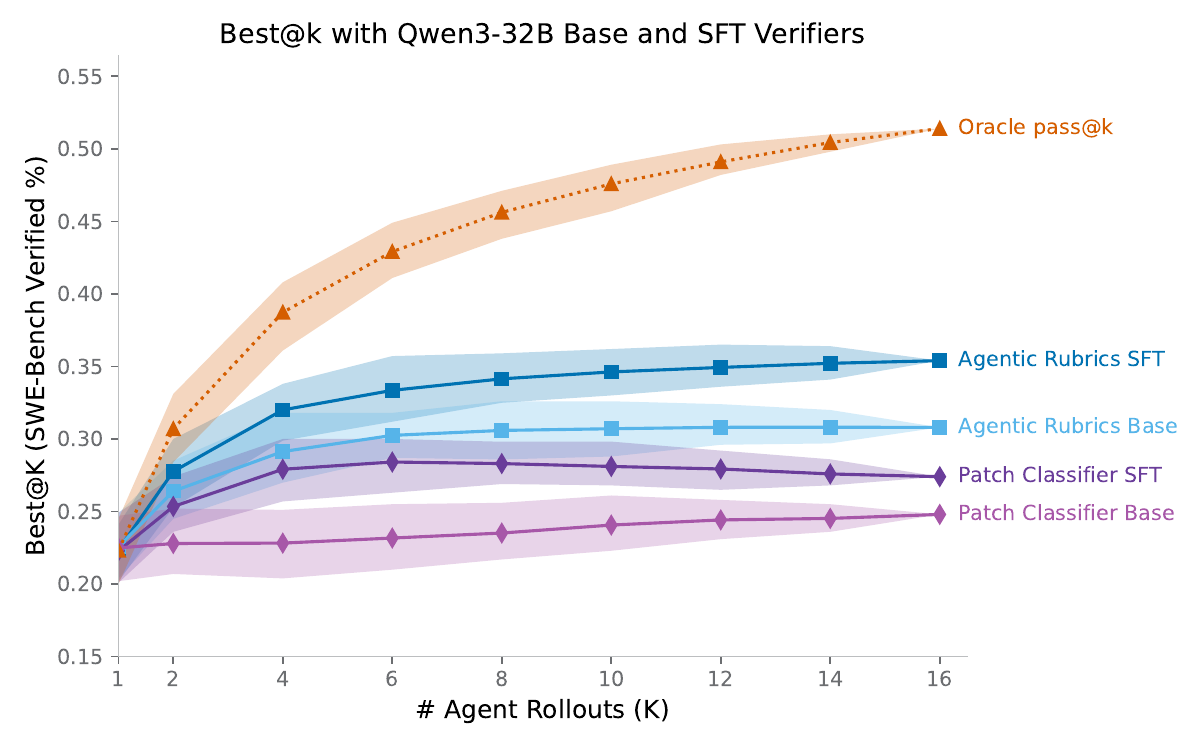}
        \caption{Finetuning (SFT) open-weight models like \texttt{Qwen3-32B} as Agentic Rubric Generator outperforms finetuning them as Patch Classifier for SWE verification. }
        \label{fig:verifier_finetuning}
    \end{figure}
\fi

\paragraph{Training Setup.}
Following \citet{jain2025r2e}, for the \textbf{patch classifier}, we fine-tune \texttt{Qwen3-32B} to output a \texttt{YES/NO} judgment given a problem statement and candidate patch. We sample 2,000 prompts from R2E-Gym and collect 4,696 test-labeled examples (approximately balanced) drawn from both expert (\texttt{Sonnet-4.5}) and on-policy (\texttt{Qwen3-32B}) rollouts. During verification, we extract the \texttt{YES/NO} token probability as the score for the patch. 
For the \textbf{agentic rubric generator}, we fine-tune \texttt{Qwen3-32B} use the agentic harness and emit a rubric file, using 2,000 rubric-generation trajectories produced by \texttt{Sonnet-4.5} (no on-policy rubric samples).
Similar to \citet{jain2025r2e}, we use the AdamW optimization for 2 epochs, a $1.0e-5$ learning rate with cosine scheduling and a batch size of 32, over 4 nodes of 8xH100 GPUs.

\paragraph{Results.}
Figure~\ref{fig:verifier_finetuning} shows that the agentic rubric generator substantially outperforms the patch-classifier verifier, and also the non-finetuned base models. This indicates the ability to produce structured, context-grounded rubrics is trainable, and is a stronger and more robust objective than binary classification for execution-free verification.


\subsection{Impact of Repository Grounding}
\label{subsec:rubric_context_gathering}

\begin{table*}[t]
\centering
\scriptsize
\begin{tabular}{p{5cm}|p{5.2cm}|p{4.4cm}}
\toprule
\textbf{Non-Agentic Rubric} & \textbf{Agentic Rubric} & \textbf{Tool Calls} \\
\midrule

\textit{``Targets code paths handling relational operators in the parser without touching unrelated operators''} &
\textit{``Adds a \texttt{visit\_Compare} method to \texttt{EvaluateFalseTransformer} class''} &
- \texttt{find -path "*/parsing/*"}\newline
- \texttt{str\_replace\_editor view}\newline
- \texttt{view --view\_range 1090 1194} \\
\midrule

\textit{``Modifies the kbd role implementation file that contains HTML generation logic''} &
\textit{``Modifies \texttt{KeyboardTransform} class in \texttt{transforms.py}''} &
- \texttt{find -exec grep -l "kbd"}\newline
- \texttt{grep -r "kbd" transforms.py}\newline
- \texttt{str\_replace\_editor view} \\



\bottomrule
\end{tabular}
\caption{Comparison of Non-Agentic vs Agentic Rubrics with the agent's tool calls that gather relevant context}
\label{tab:agent-agentless-rubric-comparison}
\end{table*}

To isolate the value of repository interaction, we compare \textbf{Agentic Rubrics} to \textbf{Non-Agentic Rubrics} generated by the same model from the problem statement alone, without access to the agentic harness or codebase (prompt in Appendix~\ref{sec:agent-prompts}). As shown in Table~\ref{tab:agent-agentless-rubric-comparison}, agentic rubrics use targeted tool calls (search and file inspection) to ground criteria in concrete repository entities (files, classes, methods), making items more specific and consistently gradable; non-agentic rubrics are often high-level (e.g., “touches the right code path”), which increases ambiguity and can lead to false positives. Empirically, on \textbf{SWE-Bench Verified}, using \texttt{Sonnet-4.5} for rubric generation without repository access reduces \textsc{Best@16} by \textbf{4.0} points on \texttt{Qwen3-32B} rollouts and \textbf{1.4} points on \texttt{Qwen3-Coder-30B-A3B} rollouts, showing that agentic context gathering improves both rubric quality and downstream selection.



\subsection{Sensitivity to Judge Model Choice}
\label{subsec:judge_comparison}
In table \ref{tab:judge_comparison}, we analyze how the capability of the judge model affects rubric grading on \texttt{Sonnet-4.5} rubrics for \texttt{Qwen3-32B} rollouts. We use three increasing reasoning efforts on the \texttt{GPT-5} model. We find that judge model capability has a small but non-trivial effect on performance. We don't require high reasoning efforts from our judge models, since rubrics are designed to be self-contained and atomic for easy grading. We also measure the flakiness of rubric grading in \ref{app:flakiness}. 

\ifCAMERA
\else
    \begin{table}[t]
    \centering
    \begin{tabular}{lc}
    \toprule
    \textbf{Judge Model} & \textbf{Best@16} \\
    \midrule
    GPT-5-mini             & 52.6 $\pm$ 2.20 \\
    GPT-5 Low Reasoning    & 54.2 $\pm$ 2.22 \\
    GPT-5 Medium Reasoning & 54.3 $\pm$ 2.25 \\
    GPT-5 High Reasoning   & 55.0 $\pm$ 2.21 \\
    \bottomrule
    \end{tabular}
    \caption{Best@16 accuracy for different judge model capabilities for scoring \texttt{Sonnet-4.5} rubrics on \texttt{Qwen3-Coder-30B-A3B} rollouts.}
    \label{tab:judge_comparison}
    \end{table}
\fi

\section{Related Work}
\paragraph{Coding Agents and Test Time Scaling}
Real world coding for SWE problems have become a popular domain for applying LLMs. SWE enviornments from open-source GitHub repos provide a good testbed for training \citep{jain2025r2e,jimenez2023swe, pan2024swegym, wei2025swerl, luodeepswe} and evaluation of coding agents \citep{openai2024swebenchverified,deng2025swebenchpro}. Agentic scaffolds like Agentless, SWEAgent, Mini SWEAgent and OpenHands define a standardized interface for using these models on such tasks \citep{xia2024agentlessdemystifyingllmbasedsoftware, yang2024swe,minisweagent2024,wang2024openhands}. Test-Time Scaling (TTS) is a way to leverage inference-time compute to improve performance on verifiable tasks like SWE Agents. \citep{ yao2023treeofthoughts, wang2023selfconsistency, brown2024largelanguagemonkeysscaling, zhu2025scalingagents}. In addition, some works study the use of verifiers for TTS and RL, but they are limited to training a reward/scoring model like a patch classifier or a testing agent \citep{pan2024swegym,jain2025r2e,luodeepswe, wei2025trainingsuperintelligentsoftwareagents}. 

\paragraph{Rubrics as verifiers for LLMs}

Rubrics have become the predominant way to evaluate LLMs on several key capabilities \citep{arora2025healthbench,akyurek2025prbench}. They have also been used as a reward signal to train LLMs during RL \citep{gunjal2025rubricsasrewards,viswanathan2025checklists, goel2025trainingaicoscientistsusing}. In this work, we describe how context-aware rubrics are an effective verifier that can holistically verify candidates for SWE tasks, and demonstrate their value through TTS.

\section{Conclusion}

Automatic, high-quality verification is essential for improving SWE agents. We study \textbf{Agentic Rubrics}, a context-grounded yet execution-free verification signal for SWE patches. Under the standard parallel test-time scaling setting on SWE-Bench Verified, Agentic Rubrics consistently outperform strong non-agentic and agentic baselines. Beyond selection performance, rubrics provide interpretable natural-language feedback and are well-aligned with human-written ground-truth tests and reference patches, while also surfacing failure modes that the available tests may not capture. Finally, our ablations study key design choices in the agentic rubric pipeline, including the role of repository interaction, the rubric agent, and the judge model. We hope these findings motivate future work on improving rubric quality and integrating rubrics as reward signals for scalable post-training of SWE agents.

\section{Limitations} 
Agentic Rubrics provide an interpretable, codebase-grounded verification signal that can be applied \emph{without test execution} once the rubric is produced. In this paper, we study them in the parallel test-time scaling setting, which offers a clean and widely used way to evaluate verifiers by holding the generator fixed and varying only the selection rule. A natural next step is to integrate rubric signals into \emph{post-training} pipelines, where they could act as rewards for RLVR-style optimization. This introduces additional challenges such as reward hacking, non-stationarity as policies improve, and credit assignment across multi-step agent behavior, which we leave as future work.

Rubric quality is another important axis. While our utility analysis shows that most automatically generated rubric judgments are \emph{high-utility} and substantively grounded, a subset of rubrics fall into low-utility modes (e.g., over-specification, redundancy, or rubric--test mismatches). This motivates \emph{human-in-the-loop} refinement as a practical next step: lightweight review/editing, rubric-template reuse, and targeted prompts for common failure modes could further improve rubric fidelity while preserving auditability, and can also generate supervision for stronger rubric-generation models.


\bibliographystyle{abbrvnat}
\bibliography{custom}
\appendix

\newpage
\section{Appendix}
\label{sec:appendix}

\subsection{Analyzing agentic rubric scores against Ground-Truth patch}

\label{app:rubric_scores_reference}

\begin{table}[h]
\centering
\small
\caption{Average Weighted Scores by Agentic Rubrics produced by different models on human written Ground-Truth Patches}
\label{tab:rubric_scores_reference}
\begin{tabular}{l c}
\toprule
\textbf{Model} & \textbf{Avg Weighted Score} \\
\midrule
Opus-4.5     & 0.8658 \\
GPT-5        & 0.8413 \\
Sonnet-4.5   & 0.8233 \\
Gemini-3-Pro & 0.8082 \\
Meta-CWM     & 0.8015 \\
Qwen3-Coder-30B-A3B  & 0.8037 \\
Qwen3-32B    & 0.6729 \\
\bottomrule
\end{tabular}
\end{table}

\begin{figure*}[t]
    \centering
    \includegraphics[width=\textwidth]{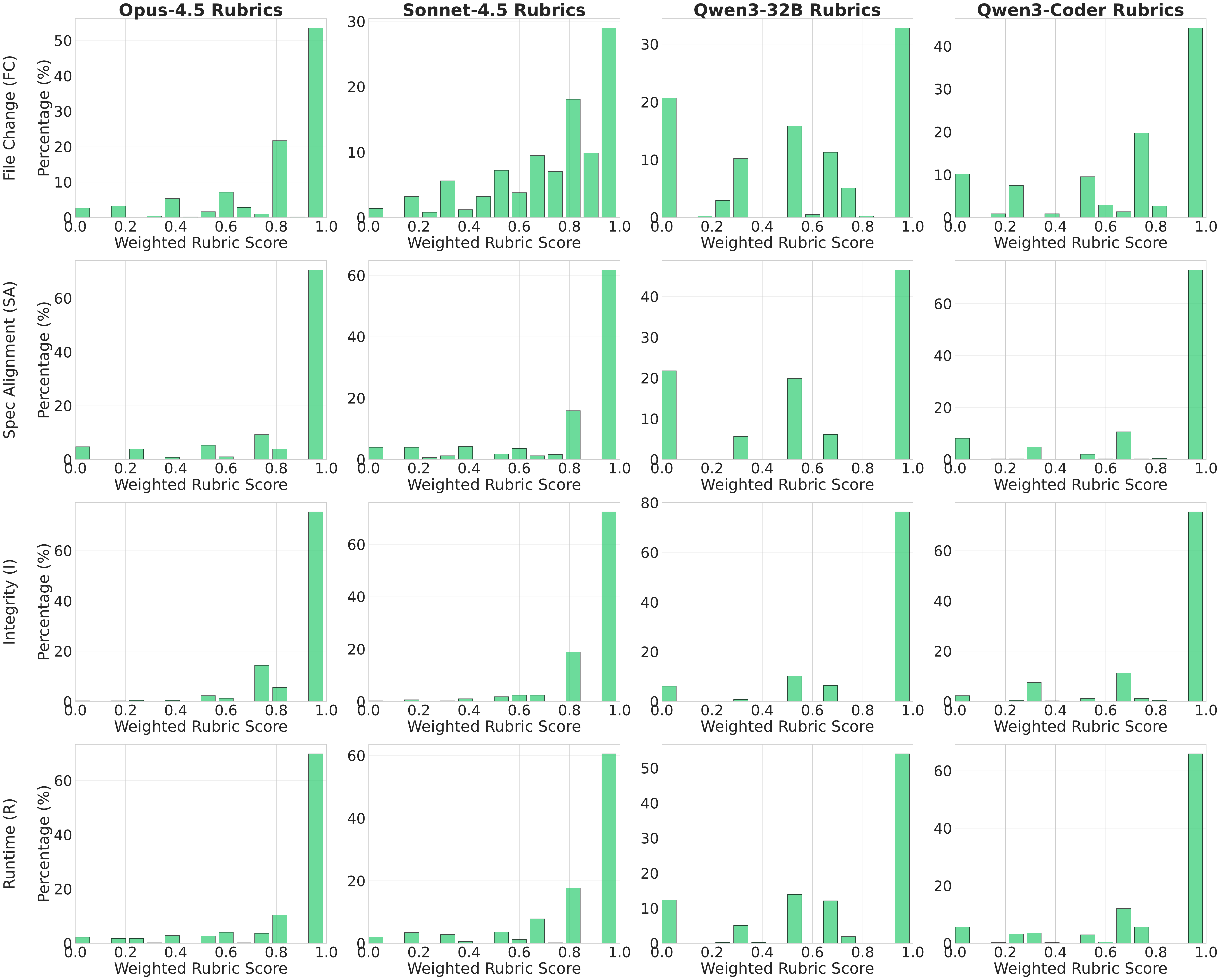}
    \caption{Distribution of rubric scores on reference patches comparing good vs bad models}
    \label{fig:reference_patch_rubric_distribution}
\end{figure*}

In table \ref{tab:rubric_scores_reference}, we show the average scores by rubrics generated by different models over the Ground-Truth patches for the tasks. We also show the breakdown across different rubric axis for a representative subset in \ref{fig:reference_patch_rubric_distribution}. Frontier Coding Models have higher alignment with human-written Ground-Truth patches than open-weight models. 

\subsection{Agentic abilities of rubric generation models}

\label{app:agent_scaffold_usage}

\ifCAMERA
\else
    \begin{figure}
        \centering
        \includegraphics[width=0.6\linewidth]{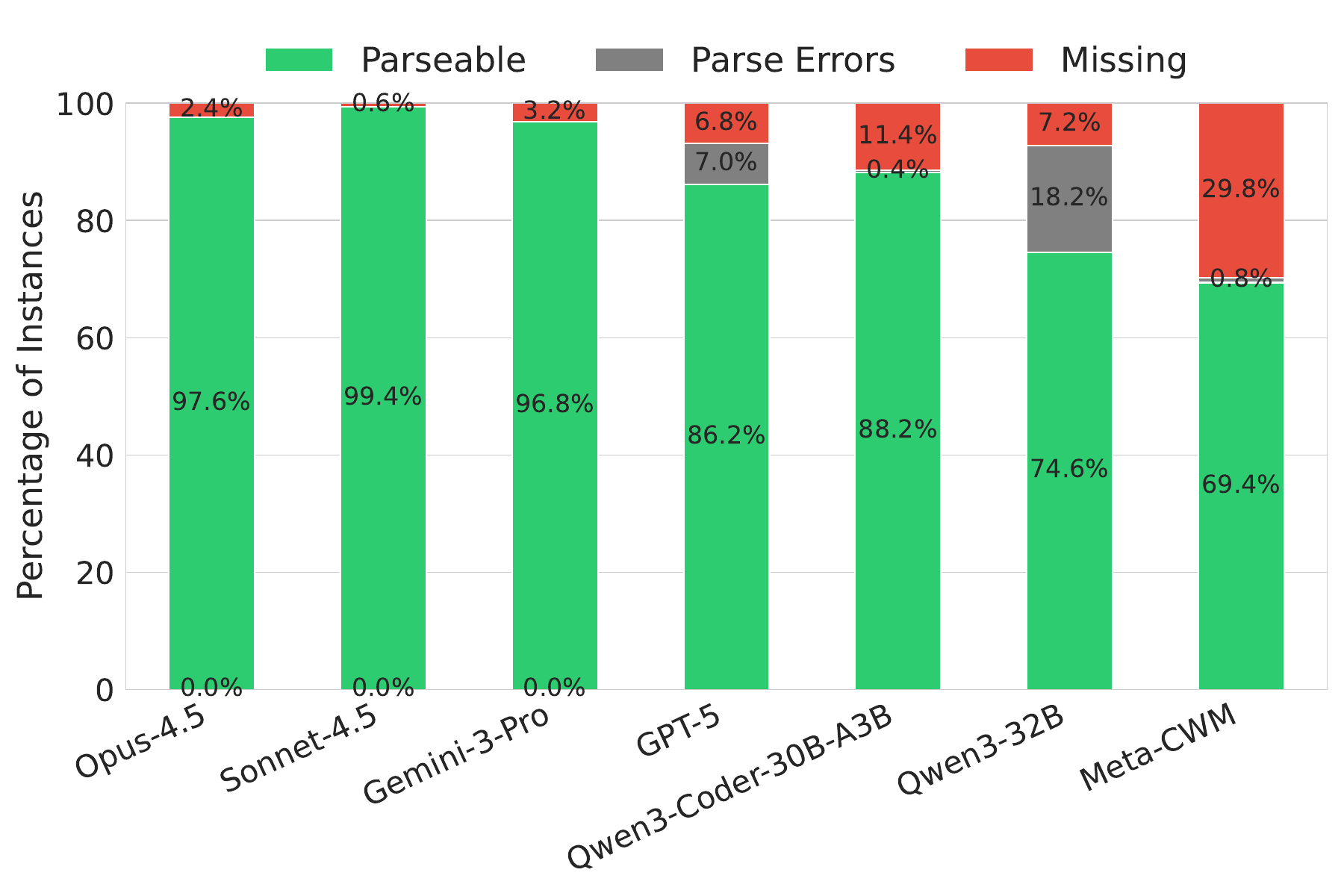}
        \caption{Performance of different models in using the Rubric Generation scaffold to create parseable rubric files.}
        \label{fig:rubric_availability_models}
    \end{figure}
\fi

Figure \ref{fig:rubric_availability_models} shows the percentage of instances where each model successfully produces parseable YAML rubric files. Frontier models demonstrate near-perfect adherence to the structured output format: Sonnet-4.5 and Gemini-3-Pro generate well-formatted rubric files 97.8\% and 96.8\% respectively, with zero parse errors. In contrast, smaller models exhibit degraded format compliance, due to weaker tool calling and instruction following. Qwen3-32B produces valid rubrics for only 74.6\% of instances with an 18.2\% parse error rate, while Meta-CWM succeeds on 69.4\% of instances but fails to generate rubrics entirely for 29.8\% of cases.

\begin{figure*}[h]
    \centering

    \begin{subfigure}[b]{0.48\textwidth}
        \centering
        \includegraphics[width=\linewidth]{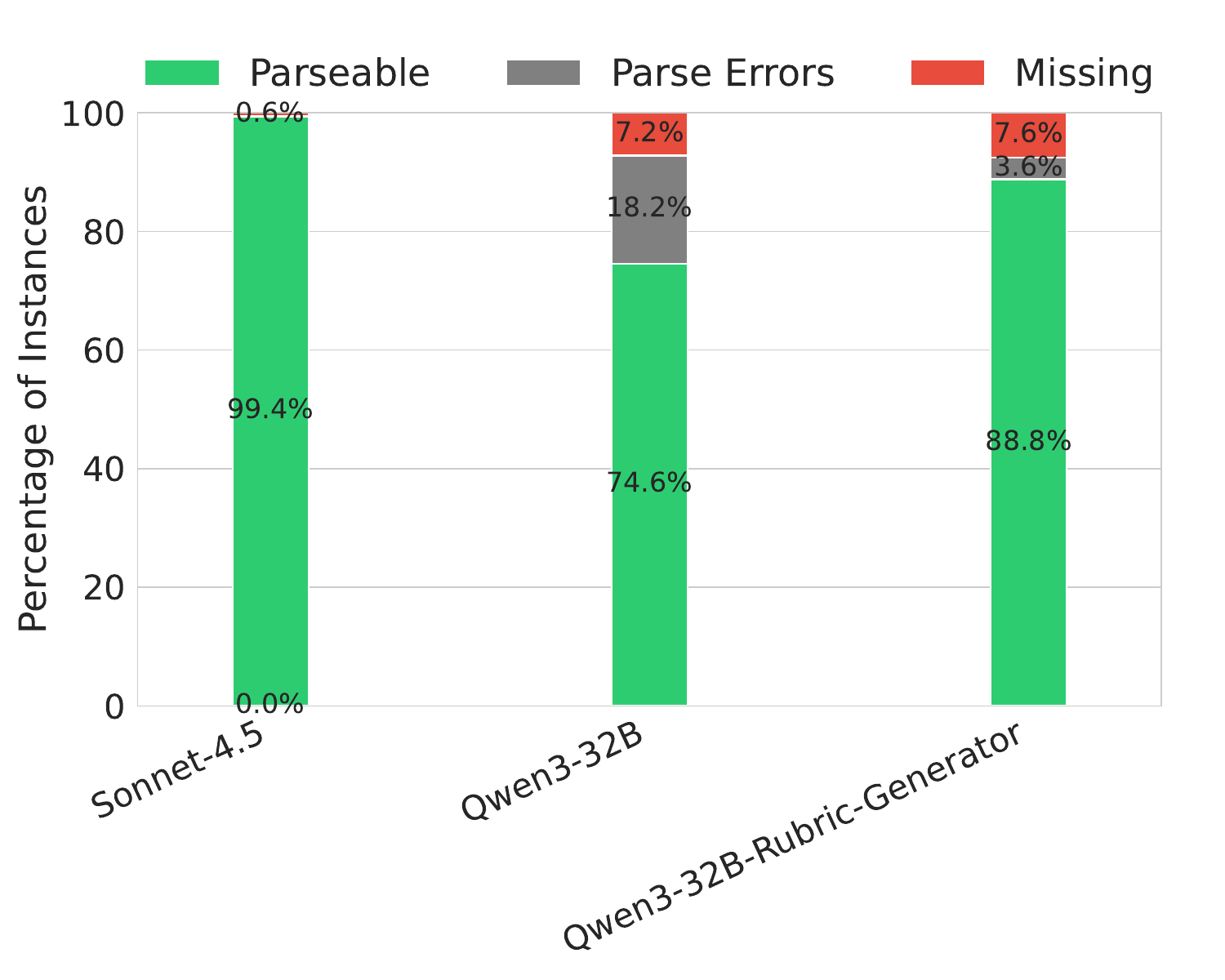}
        \caption{Agentic harness use improvement through finetuning}
        \label{fig:rubric_availability_models_finetuned}
    \end{subfigure}
    \hfill
    \begin{subfigure}[b]{0.48\textwidth}
        \centering
        \includegraphics[width=\linewidth]{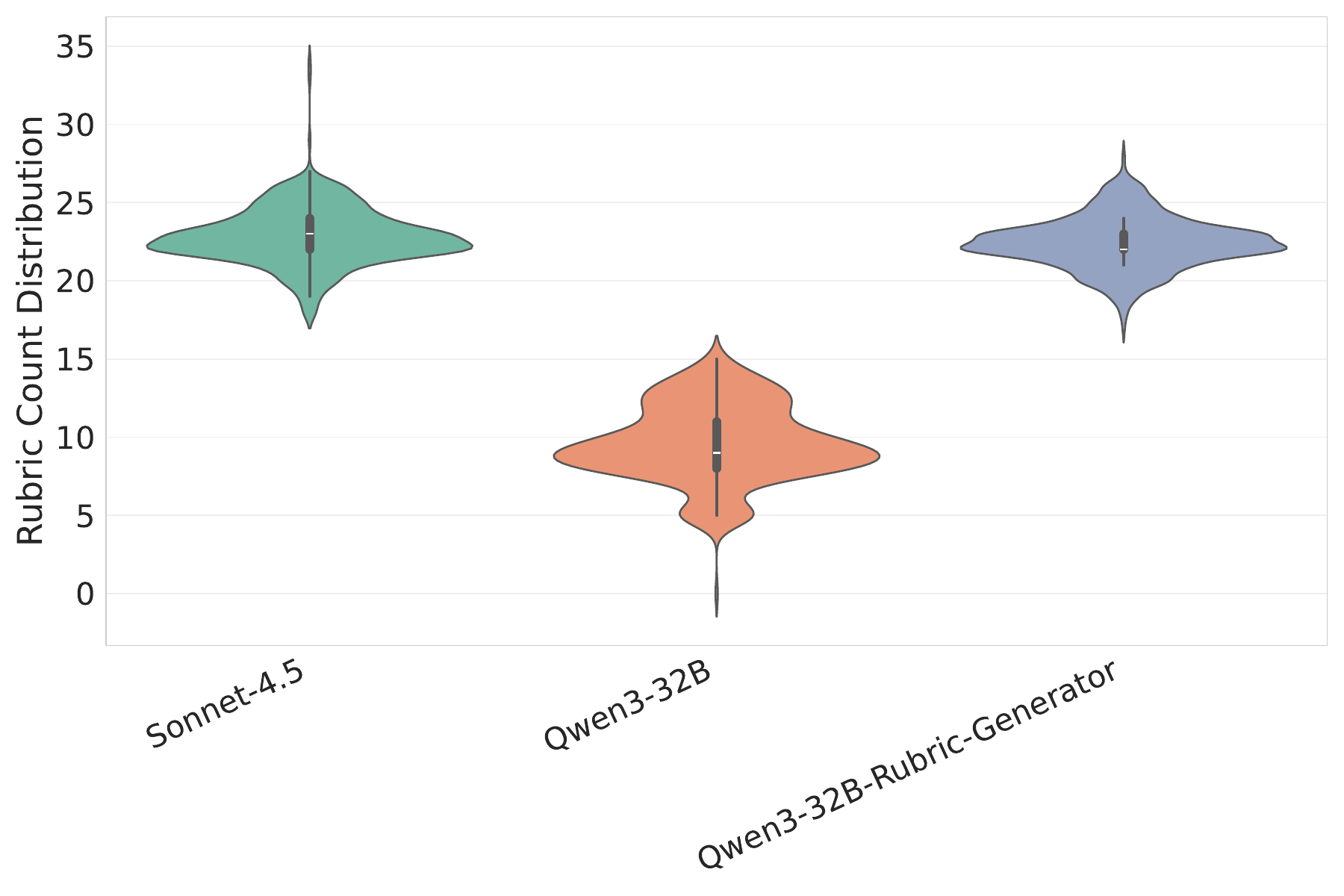}
        \caption{Rubric structure improvement through finetuning}
        \label{fig:avg_rubric_distribution_finetuned}
    \end{subfigure}
    
    \caption{Finetuning Qwen3-32B on Sonnet-4.5 rubric agent trajectories leads to (a) better use of the Agentic harness as demonstrated by reduced errors and (b) improved rubric distribution, matching the original model.}
    \label{fig:rubric_finetuned}
\end{figure*}

In figure \ref{fig:rubric_finetuned}, we show how finetuning Qwen3-32B on Sonnet-4.5 rubric agent trajectories leads to better use of the Agentic harness as demonstrated by reduced errors and improved rubric distribution, matching the original teacher model. This demonstrates that we can train rubric generator models, which unlocks their use for creating preference data and reward models.

\subsection{Cost analysis for agentic verification methods}

\begin{table}[!h]
\centering
\small
\begin{tabular}{lcccc}
\toprule
\multirow{2}{*}{\textbf{Method}} & \textbf{Artifact} & \textbf{Grading} & \textbf{API} & \textbf{Qwen3-32B} \\
& \textbf{Cost(\$)} & \textbf{Cost(\$)} & \textbf{Calls} & \textbf{Best@16} \\
\midrule
Patch Sim. & 0.640 & 0.006 & 48.5 & 36.6 \\
Tests & 0.499 & 0.001 & 29.2 & 33.6 \\
Rubrics & 0.245 & 0.003 & 22.9 & 40.6 \\
\bottomrule
\end{tabular}
\caption{Cost vs Performance (Qwen3-32B Best@K) comparison of different agentic verifier methods that all use the same underlying model, Sonnet-4.5. Artifact Cost is the average cost in USD (\$) for the total input and output tokens in the entire trajectory. Note that agentic patch generation was run with 50 steps as the limit to improve solution patch quality since the reference patches produced with the 30 turn limit was too restrictive to get any comparable result, while the other methods are limited to 30 turns.}
\label{tab:cost-performance}
\end{table}

In table \ref{tab:cost-performance}, we describe the cost analysis of different agentic methods. Note that the cost of Grading for Tests is spinning up a sandboxed environment for grading and running the test suite after applying the patch. We use Modal to run sandboxed containers. All costs are averaged over the entire dataset. The cost of artifact is once per instance (a fixed cost) while cost of grading per rollout, and will scale with the number of rollouts used. Therefore the average total cost of \textsc{BEST@16} calculation  (Agentic artifact generation + grading 16 rollouts) per instance is \$0.736 for Patch Similarity, \$0.515 for Test Generation and \$0.293 for Rubrics.

We see that rubric generation is very cost efficient, requiring fewer tool calls and tokens while also achieving higher performance.

\subsection{Rubric Flakiness Study}
\label{app:flakiness}

To assess the reliability of LLM-based rubric evaluation, we do a flakiness study measuring grading determinism across repeated trials. We randomly sampled 20 instances for rubric generator models \texttt{Sonnet-4.5}, and \texttt{Qwen3-32B}, selecting 5 rubric items per instance (100 items per model). Each rubric item was then scored 5 independent times using \texttt{GPT-5 (low reasoning)} as the judge, producing binary assessments of whether the candidate patch satisfied the rubric criterion.

A rubric item is considered ``flaky'' if any of its 5 trial scores differ from the others. Our results demonstrate high scoring determinism, and stronger models write better, less flaky rubrics: \texttt{Sonnet-4.5} generated rubrics exhibited only 2\% flakiness (98\% of items scored identically across all trials), while \texttt{Qwen3-32B} had 9\% flakiness. We attribute such low levels of flakiness to our instructions that mandate rubrics to be atomic and self-contained, reducing the scope for the Judge's interpretation for grading.  

High consistency rates are important, since they lead to reproducible assessments and lower gaming opportunities. Future work can study this further, to establish best practices for writing strong, deterministic yet non-prescriptive rubrics.

\subsection{Hybrid verifiers using rubrics v/s classifier}

\ifCAMERA
\else
    \begin{figure}[h]
    \centering
    \includegraphics[width=0.6\linewidth]{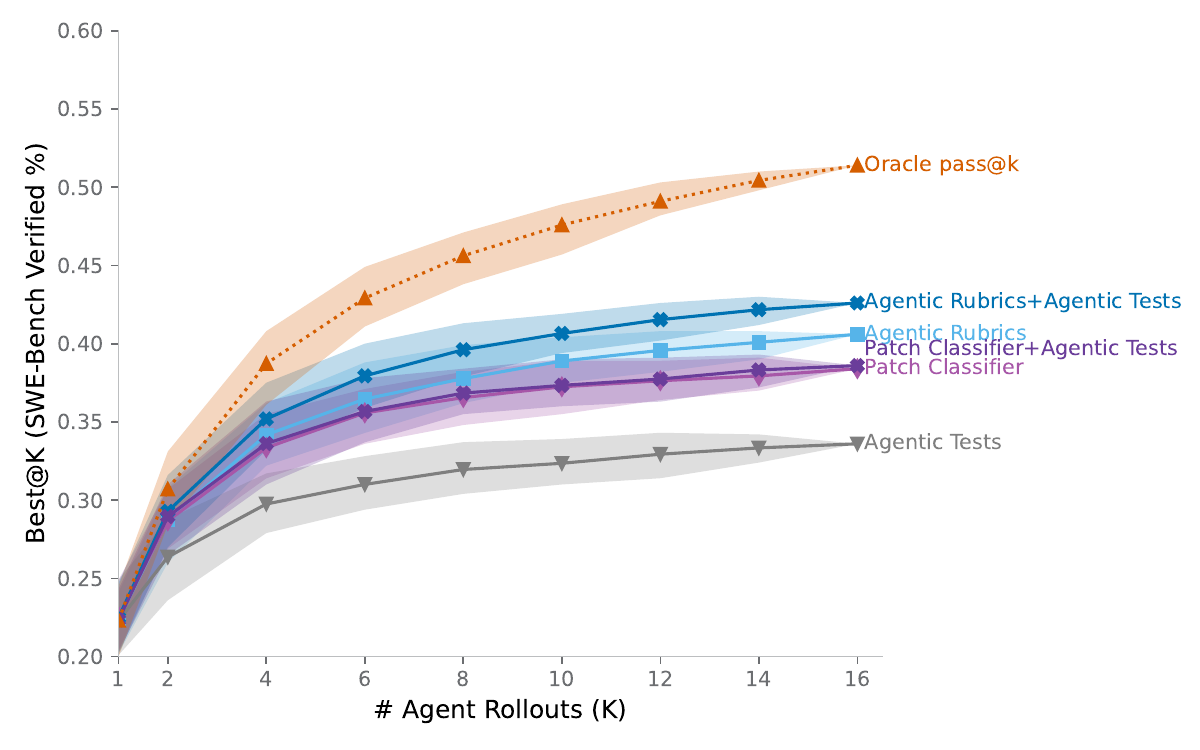}
    \caption{Combining Agentic Rubrics with Agentic Tests, to build a Hybrid Verifier.}
    \label{fig:hybrid_methods}
\end{figure}
\fi 

We also study how agentic rubrics compare against a classifier based approach when combined with generated tests in a hybrid approach similar to R2E-Gym \citep{jain2025r2e}. In figure \ref{fig:hybrid_methods}, we see that a hybrid verifier setup where we combine agentic tests and rubrics in a simple aggregation setup can work better than both methods in isolation. This opens up future work in combining these verifiers in more complex pipelines to extract maximum utility from different verfication methods.  

\subsection{Categories of rubric utility classification}

In table \ref{tab:rubric_taxonomy}, we describe the taxonomy used to label rubric utility. This was crafted manually by inspecting the rollouts and failure modes with different rubrics. 

\begin{table*}[h]
\centering
\small
\begin{tabularx}{\linewidth}{l l X}
\toprule
\textbf{Tier} & \textbf{Sub-category} & \textbf{Description} \\
\midrule
High-Utility &
Rule Break &
Violation of an internal contract or assumption that should hold even if tests do not exercise it directly. \\

High-Utility &
Band-Aid Fix &
Patch makes tests pass but does not actually fix the underlying semantics or root cause of the bug. \\

High-Utility &
Wrong Layer &
Fix is implemented in the wrong module, class, or layer; the behavior change should live elsewhere in the codebase. \\

High-Utility &
Root Cause Missed &
Core bug remains present or is only partially addressed; the patch does not truly resolve the reported issue. \\

High-Utility &
Missing Edges &
Important edge cases or reproducer scenarios implied by the spec are not handled, even though the main path may pass tests. \\

High-Utility &
Scope Creep &
Patch introduces unrelated or off-scope changes (e.g., extra files, debug code, refactors, binary blobs). \\

High-Utility &
Perf Risk &
Patch is likely to introduce a non-trivial performance or resource-usage regression that tests do not directly measure. \\

High-Utility &
API Break &
Patch breaks a public API or its backward-compatible behavior as relied on by existing callers or downstream code. \\

High-Utility &
Security Risk &
Patch weakens validation, safety, or security properties (e.g., injection, data exposure) beyond what tests explicitly guard. \\
\midrule
Low-Utility &
Test Rules Mismatch &
Rubric assumes a different test setup or evaluation protocol than the one actually used (e.g., expects modifying tests or harness behavior). \\

Low-Utility &
Over-Specified Fix &
Rubric demands a particular implementation strategy or pattern even though multiple correct fixes are allowed by the spec. \\

Low-Utility &
API Over-Strict &
Rubric penalizes benign API or structural changes even when observable behavior matches the ground-truth spec. \\

Low-Utility &
Style Nit &
Rubric focuses on purely stylistic or cosmetic issues with no semantic impact on correctness or behavior. \\

Low-Utility &
Spec Clash &
Rubric contradicts the problem statement, golden patch, or golden test cases (e.g., forbids behavior that the reference explicitly permits). \\

Low-Utility &
Ref Patch Conflict &
Golden patch itself violates the rubric’s stated constraint, indicating that the rubric is misaligned with the reference solution. \\

Low-Utility &
Redundant Signal &
Rubric adds no new information beyond other rubrics, effectively double-counting the same issue without additional insight. \\

Low-Utility &
Eval Bug &
Rubric failure arises from a scoring, matching, or parsing bug rather than an actual violation of the rubric text by the candidate patch. \\

Low-Utility &
Irrelevant Rule &
Rubric encodes a constraint that is not actually needed or justified for this particular bug or problem context. \\
\bottomrule
\end{tabularx}
\caption{Rubric category taxonomy used in utility analysis.}
\label{tab:rubric_taxonomy}
\end{table*}

\subsection{SWE Agent Setup}

We use the SWE-Agent scaffold \cite{yang2024swe} for all agentic setups. We add a turn reminder every 5 turns about the number of turns left for the agent before autosubmission. In addition, we wanted to study and distill rubric generation capability but many state-of-the-art models completely skip any thinking or assistant tokens and just returns the tool call with the default SWE-Agent setup. So we change the parsing function to  explicitly mandate that in each turn, it return a short summary of its thinking. Prompts for all methods can be found in \ref{sec:agent-prompts}.

\subsection{Rubric and their grading - Illustrative Examples}

In \ref{app:rubric_qualitative_examples} we show three representative
examples in which the patches pass all the Ground-Truth tests for the problem but still score low on rubrics. In particular, in the first example of \texttt{matplotlib\_\_matplotlib-26291}, users report that creating an inset axis and saving a figure with \texttt{bbox\_inches="tight"} crashes with an
\texttt{AttributeError}. The root cause is that Matplotlib calls a locator object with \texttt{renderer = None}, while downstream code assumes a valid renderer.

The \emph{candidate patch} from Qwen3-32B guards against this by returning a
dummy zero-size bounding box whenever the renderer is missing or an error
occurs. This is enough to satisfy the ground-truth tests.

However, the \emph{rubrics} encode a stronger notion of correctness. They require that the fix
(1) preserves the positioning and sizing of inset axes (R4), (2) works for
both absolute and relative size specifications(SA4), and (3) still works in the original failure mode with \texttt{bbox\_inches='tight'} (SA5), without
breaking existing user code (R3). These checks treat the inset axis as a real visual element whose layout must remain meaningful. The candidate's dummy box violates this implicit contract: it causes tight layout to ignore
the inset entirely. The user no longer sees a crash, but also no longer gets
a correctly placed inset in the ``tight'' saved figure.

\onecolumn
\subsection{Rubric Examples and grading}

\label{app:rubric_qualitative_examples}

\begin{tcolorbox}[
    enhanced,
    attach boxed title to top left={xshift=6mm,yshift=-2mm},
    colback=gray!5,
    colframe=gray!70,
    colbacktitle=gray!50,
    title=matplotlib\_\_matplotlib-26291,
    fonttitle=\bfseries\color{black},
    boxed title style={size=small,colframe=gray,sharp corners},
    sharp corners,
    breakable
]
\scriptsize

\vspace{0.3em}
\textbf{Problem.} Creating inset axes and saving with
\texttt{bbox\_inches='tight'} can raise
\texttt{AttributeError: 'NoneType' object has no attribute '\_get\_renderer'}
because the inset locator is called with \texttt{renderer=None} and attempts
to access the figure renderer through a \texttt{None} figure.

\vspace{0.6em}
\textbf{Candidate patch (core diff)}
\begin{lstlisting}[
    basicstyle=\ttfamily\scriptsize,
    breaklines=true,
    columns=fullflexible,
    keepspaces=true,
]
--- a/lib/mpl_toolkits/axes_grid1/inset_locator.py
+++ b/lib/mpl_toolkits/axes_grid1/inset_locator.py
@@ -70,7 +70,18 @@ class AnchoredLocatorBase(AnchoredOffsetbox):
 
     def __call__(self, ax, renderer):
         self.axes = ax
-        bbox = self.get_window_extent(renderer)
+        if renderer is None:
+            # No renderer available, return a dummy bounding box of zero size
+            # to avoid errors during tight layout.
+            dummy_bbox = Bbox.from_bounds(0, 0, 0, 0)
+            tr = ax.figure.transSubfigure.inverted()
+            return TransformedBbox(dummy_bbox, tr)
+        try:
+            bbox = self.get_window_extent(renderer)
+        except (AttributeError, RuntimeError) as e:
+            # Handle cases where the renderer is not available, returning
+            # a dummy bbox again.
+            dummy_bbox = Bbox.from_bounds(0, 0, 0, 0)
+            tr = ax.figure.transSubfigure.inverted()
+            return TransformedBbox(dummy_bbox, tr)
         px, py = self.get_offset(bbox.width, bbox.height, 0, 0, renderer)
         bbox_canvas = Bbox.from_bounds(px, py, bbox.width, bbox.height)
         tr = ax.figure.transSubfigure.inverted()
         return TransformedBbox(bbox_canvas, tr)
\end{lstlisting}

\vspace{0.6em}

\textbf{Rubrics (\textcolor{red!70!black}{FAILED} / \textcolor{green!70!black}{PASSED} for candidate patch)}

\textcolor{red!70!black}{- FC1: Modifies the \texttt{AnchoredLocatorBase.\_\_call\_\_} method in \texttt{lib/mpl\_toolkits/axes\_grid1/inset\_locator.py} to set \texttt{self.figure} before calling \texttt{get\_window\_extent}.}\\
\textcolor{green!50!black}{- FC2: Ensures the fix is localized to the \texttt{\_\_call\_\_} method in \texttt{AnchoredLocatorBase} without modifying unrelated methods or classes.}\\
\textcolor{red!70!black}{- FC3: Sets \texttt{self.figure} from \texttt{ax.figure} in \texttt{AnchoredLocatorBase.\_\_call\_\_} before the \texttt{get\_window\_extent} call.}\\
\textcolor{green!50!black}{- FC4: Avoids modifying the parent class \texttt{AnchoredOffsetbox} in \texttt{lib/matplotlib/offsetbox.py} or its \texttt{get\_window\_extent} method.}\\
\textcolor{green!50!black}{- FC5: Does not introduce changes to other locator classes like \texttt{AnchoredSizeLocator} or \texttt{AnchoredZoomLocator} beyond what is inherited from \texttt{AnchoredLocatorBase}.}\\
\textcolor{green!50!black}{- FC6: Avoids adding unnecessary imports or dependencies to \texttt{inset\_locator.py}.}\\[0.3em]

\textcolor{green!50!black}{- I1: Does not weaken or skip existing tests in \texttt{lib/mpl\_toolkits/axes\_grid1/tests/test\_axes\_grid1.py} that verify \texttt{inset\_axes} functionality.}\\
\textcolor{green!50!black}{- I2: Maintains the public API signature of \texttt{inset\_axes} without adding or removing parameters.}\\
\textcolor{green!50!black}{- I3: Preserves the existing behavior of \texttt{AnchoredLocatorBase.\_\_call\_\_} returning a \texttt{TransformedBbox} as before.}\\
\textcolor{green!50!black}{- I4: Does not modify the constructor \texttt{\_\_init\_\_} of \texttt{AnchoredLocatorBase} or its parent class initialization.}\\
\textcolor{green!50!black}{- I5: Avoids introducing deprecation warnings or changing the inheritance hierarchy of \texttt{AnchoredLocatorBase}.}\\[0.3em]

\textcolor{red!70!black}{- R1: Ensures \texttt{self.figure} is set to a non-\texttt{None} \texttt{Figure} instance before \texttt{get\_window\_extent} accesses \texttt{self.figure.\_get\_renderer()}.}\\
\textcolor{green!50!black}{- R2: Allows \texttt{inset\_axes} creation to proceed through the full render path, including \texttt{tight\_bbox.adjust\_bbox}, without \texttt{AttributeError}.}\\
\textcolor{red!70!black}{- R3: Maintains backward compatibility so existing code using \texttt{inset\_axes} continues to work without modification.}\\
\textcolor{red!70!black}{- R4: Preserves correct positioning and sizing of inset axes relative to the parent axes after the fix.}\\
\textcolor{green!50!black}{- R5: Ensures the fix handles both cases where \texttt{renderer} is \texttt{None} and where \texttt{renderer} is explicitly passed to \texttt{get\_window\_extent}.}\\
\textcolor{green!50!black}{- R6: Allows the locator to be called multiple times on the same axes without error or state corruption.}\\[0.3em]

\textcolor{green!50!black}{- SA1: Resolves the \texttt{AttributeError: 'NoneType' object has no attribute '\_get\_renderer'} that occurs when calling \texttt{inset\_axes} with the code example from the bug report.}\\
\textcolor{red!70!black}{- SA2: Allows \texttt{get\_window\_extent} to successfully access \texttt{self.figure.\_get\_renderer()} when \texttt{renderer} is \texttt{None} in the \texttt{offsetbox.py} implementation.}\\
\textcolor{green!50!black}{- SA3: Preserves the existing behavior where \texttt{self.axes} is set to the \texttt{ax} parameter in \texttt{AnchoredLocatorBase.\_\_call\_\_}.}\\
\textcolor{red!70!black}{- SA4: Maintains compatibility with both absolute size specifications like \texttt{width=1.3} and relative specifications like \texttt{width='30\%'} as shown in the \texttt{inset\_locator\_demo.py} examples.}\\
\textcolor{red!70!black}{- SA5: Ensures the fix works when \texttt{bbox\_inches='tight'} is used during figure saving, which triggers the \texttt{adjust\_bbox} path that calls the locator.}\\

\vspace{0.3em}
\end{tcolorbox}



\begin{tcolorbox}[
    enhanced,
    attach boxed title to top left={xshift=6mm,yshift=-2mm},
    colback=gray!5,
    colframe=gray!70,
    colbacktitle=gray!50,
    title=matplotlib\_\_matplotlib-25332,
    fonttitle=\bfseries\color{black},
    boxed title style={size=small,colframe=gray,sharp corners},
    sharp corners,
    breakable
]
\scriptsize

\vspace{0.3em}
\textbf{Problem.} After calling \texttt{fig.align\_labels()} or
\texttt{fig.align\_ylabels()}, \texttt{pickle.dumps(fig)} fails with
\texttt{TypeError: cannot pickle 'weakref.ReferenceType' object} because
\texttt{Grouper} stores weak references that are not directly pickle-safe.

\vspace{0.6em}

\textbf{Candidate patch (full diff)}
\begin{lstlisting}[
    basicstyle=\ttfamily\scriptsize,
    breaklines=true,
    columns=fullflexible,
    keepspaces=true,
]
diff --git a/lib/matplotlib/cbook.py b/lib/matplotlib/cbook.py
index c9699b2e21..1adf054843 100644
--- a/lib/matplotlib/cbook.py
+++ b/lib/matplotlib/cbook.py
@@ -752,6 +752,24 @@ def print_cycles(objects, outstream=sys.stdout, show_progress=False):
 
 
 class Grouper:
+    def __getstate__(self):
+        # Convert weakrefs to strong references for pickling
+        clean_mapping = {}
+        for key, value in self._mapping.items():
+            # key is a weakref, value is a list of weakrefs
+            strong_key = key()
+            strong_values = [ref() for ref in value]
+            clean_mapping[strong_key] = strong_values
+        return clean_mapping
+
+    def __setstate__(self, clean_mapping):
+        # Reconstruct the _mapping with weakrefs from strong references
+        self._mapping = {}
+        for key, value in clean_mapping.items():
+            # key is a strong reference, value is a list of strong references
+            weak_key = weakref.ref(key)
+            weak_values = [weakref.ref(val) for val in value]
+            self._mapping[weak_key] = weak_values
     """"""
     A disjoint-set data structure.
\end{lstlisting}

\vspace{0.6em}

\textbf{Rubrics (\textcolor{red!70!black}{FAILED} / \textcolor{green!70!black}{PASSED} for candidate patch)}

\textcolor{red!70!black}{- FC1: Modifies \texttt{lib/matplotlib/figure.py} to handle \texttt{\_align\_label\_groups} during pickling in \texttt{\_\_getstate\_\_} or \texttt{\_\_setstate\_\_}.}\\
\textcolor{green!50!black}{- FC2: Avoids modifying unrelated files beyond \texttt{lib/matplotlib/figure.py}, \texttt{lib/matplotlib/cbook.py}, and \texttt{lib/matplotlib/tests/test\_pickle.py}.}\\
\textcolor{green!50!black}{- FC3: Adds or modifies pickle support methods (\texttt{\_\_getstate\_\_}, \texttt{\_\_setstate\_\_}, \texttt{\_\_reduce\_\_}, or \texttt{\_\_reduce\_ex\_\_}) specifically to handle \texttt{Grouper} weak references.}\\
\textcolor{green!50!black}{- FC4: Preserves existing pickling logic in \texttt{Figure.\_\_getstate\_\_} for canvas, \texttt{\_dpi}, \texttt{\_\_mpl\_version\_\_}, and \texttt{\_restore\_to\_pylab}.}\\
\textcolor{green!50!black}{- FC5: Implements the solution either by modifying \texttt{cbook.Grouper} class to add pickle support or by handling \texttt{\_align\_label\_groups} specially in \texttt{Figure}'s pickle methods.}\\
\textcolor{green!50!black}{- FC6: Avoids adding unnecessary complexity such as deep copying the entire figure state or serializing canvas data that should remain transient.}\\[0.3em]

\textcolor{red!70!black}{- I1: Adds a test case in \texttt{lib/matplotlib/tests/test\_pickle.py} that reproduces the exact scenario from the bug description with \texttt{align\_labels()} call.}\\
\textcolor{green!50!black}{- I2: Avoids weakening or skipping existing pickle tests in \texttt{test\_pickle.py}.}\\
\textcolor{green!50!black}{- I3: Maintains backward compatibility by not changing the public API of \texttt{Figure.align\_labels}, \texttt{Figure.align\_xlabels}, or \texttt{Figure.align\_ylabels}.}\\
\textcolor{green!50!black}{- I4: Does not break existing functionality where figures without \texttt{align\_labels()} can still be pickled successfully.}\\
\textcolor{red!70!black}{- I5: Preserves the \texttt{Grouper} API if modifying \texttt{cbook.Grouper} class, keeping \texttt{join()}, \texttt{joined()}, \texttt{get\_siblings()}, \texttt{clean()}, and \texttt{\_\_iter\_\_()} methods functional.}\\
\textcolor{green!50!black}{- I6: Avoids removing or commenting out the \texttt{\_align\_label\_groups} attribute initialization in \texttt{Figure.\_\_init\_\_}.}\\[0.3em]

\textcolor{green!50!black}{- R1: Maintains correct label alignment behavior where labels in the same row or column remain visually aligned after pickle round-trip.}\\
\textcolor{green!50!black}{- R2: Ensures the unpickled figure's \texttt{\_align\_label\_groups} contains valid \texttt{Grouper} objects or equivalent structures that preserve axis grouping relationships.}\\
\textcolor{green!50!black}{- R3: Prevents \texttt{AttributeError} or \texttt{KeyError} when accessing \texttt{\_align\_label\_groups} after unpickling the figure.}\\
\textcolor{red!70!black}{- R4: Ensures that calling \texttt{align\_labels()} again on an unpickled figure does not raise exceptions and properly updates groupings.}\\
\textcolor{green!50!black}{- R5: Preserves memory efficiency by not creating strong references to axes objects where weak references were intentionally used.}\\
\textcolor{red!70!black}{- R6: Handles edge cases where axes are deleted or garbage collected between pickling and unpickling without causing crashes.}\\[0.3em]

\textcolor{green!50!black}{- SA1: Ensures \texttt{pickle.dumps(fig)} succeeds after calling \texttt{fig.align\_labels()} without raising \texttt{TypeError} about \texttt{weakref.ReferenceType}.}\\
\textcolor{green!50!black}{- SA2: Ensures \texttt{pickle.dumps(fig)} succeeds after calling \texttt{fig.align\_xlabels()} independently.}\\
\textcolor{green!50!black}{- SA3: Ensures \texttt{pickle.dumps(fig)} succeeds after calling \texttt{fig.align\_ylabels()} independently.}\\
\textcolor{red!70!black}{- SA4: Preserves label alignment behavior after unpickling so that aligned labels remain properly grouped.}\\
\textcolor{red!70!black}{- SA5: Allows round-trip pickling where \texttt{pickle.loads(pickle.dumps(fig))} produces a functional figure with preserved alignment state.}

\vspace{0.3em}
\end{tcolorbox}

\begin{tcolorbox}[
    enhanced,
    attach boxed title to top left={xshift=6mm,yshift=-2mm},
    colback=gray!5,
    colframe=gray!70,
    colbacktitle=gray!50,
    title=django\_\_django-13417,
    fonttitle=\bfseries\color{black},
    boxed title style={size=small,colframe=gray,sharp corners},
    sharp corners,
    breakable
]
\scriptsize

\vspace{0.3em}
\textbf{Problem.} For models with \texttt{Meta.ordering}, annotated querysets
that introduce \texttt{GROUP BY} clauses (e.g., via \texttt{annotate(Count('pk'))})
still report \texttt{qs.ordered} as \texttt{True} even though the generated SQL
no longer includes an \texttt{ORDER BY}. This makes the \texttt{ordered}
property inconsistent with the actual ordering behavior.

\vspace{0.6em}
\textbf{Candidate patch (core diff)}
\begin{lstlisting}[
    basicstyle=\ttfamily\scriptsize,
    breaklines=true,
    columns=fullflexible,
    keepspaces=true,
]
diff --git a/django/db/models/sql/query.py b/django/db/models/sql/query.py
index 4648daf395..e1672b0c1c 100644
--- a/django/db/models/sql/query.py
+++ b/django/db/models/sql/query.py
@@ -1030,6 +1030,7 @@ class Query(BaseExpression):
         else:
             self.set_annotation_mask(set(self.annotation_select).difference({alias}))
         self.annotations[alias] = annotation
+        self.clear_ordering(True)
 
     def resolve_expression(self, query, *args, **kwargs):
         clone = self.clone()
@@ -1038,10 +1039,13 @@ class Query(BaseExpression):
         clone.subquery = True
         # It's safe to drop ordering if the queryset isn't using slicing,
         # distinct(*fields) or select_for_update().
+        # Clear ordering if annotation requires grouping
+        self.clear_ordering(True)
         if (self.low_mark == 0 and self.high_mark is None and
                 not self.distinct_fields and
                 not self.select_for_update):
             clone.clear_ordering(True)
+        self.clear_ordering(True)
         clone.where.resolve_expression(query, *args, **kwargs)
         for key, value in clone.annotations.items():
             resolved = value.resolve_expression(query, *args, **kwargs)
\end{lstlisting}

\vspace{0.6em}

\textbf{Rubrics (\textcolor{red!70!black}{FAILED} / \textcolor{green!70!black}{PASSED} for candidate patch)}

\textcolor{red!70!black}{- FC1: Modifies the \texttt{ordered} property in \texttt{django/db/models/query.py} to check if \texttt{query.group\_by} is set and if ordering comes only from \texttt{Meta.ordering}.}\\
\textcolor{green!50!black}{- FC2: Avoids modifying \texttt{django/db/models/sql/compiler.py} lines 590--591 where \texttt{order\_by} is set to \texttt{None} when \texttt{\_meta\_ordering} is \texttt{True} and grouping exists.}\\
\textcolor{green!50!black}{- FC3: Preserves the existing logic in \texttt{QuerySet.ordered} for checking \texttt{self.query.extra\_order\_by}, \texttt{self.query.order\_by}, and \texttt{self.query.default\_ordering}.}\\
\textcolor{red!70!black}{- FC4: Adds a check in \texttt{QuerySet.ordered} that returns \texttt{False} when \texttt{self.query.group\_by} is not \texttt{None} and ordering only comes from \texttt{self.query.get\_meta().ordering}.}\\
\textcolor{green!50!black}{- FC5: Does not modify the \texttt{Query.default\_ordering} attribute or \texttt{Query.set\_group\_by} method in \texttt{django/db/models/sql/query.py}.}\\
\textcolor{red!70!black}{- FC6: Limits changes to the \texttt{ordered} property method between lines 1218--1230 in \texttt{django/db/models/query.py} without altering other \texttt{QuerySet} methods.}\\[0.3em]

\textcolor{green!50!black}{- I1: Avoids weakening existing tests by adding \texttt{pytest.mark.skip} or removing assertions from test files.}\\
\textcolor{green!50!black}{- I2: Does not modify the public API of \texttt{QuerySet} class beyond fixing the \texttt{ordered} property behavior.}\\
\textcolor{green!50!black}{- I3: Maintains the \texttt{EmptyQuerySet} check in \texttt{ordered} that returns \texttt{True} for empty querysets.}\\
\textcolor{red!70!black}{- I4: Preserves \texttt{query.default\_ordering} attribute semantics without changing its value based on \texttt{GROUP BY} presence.}\\
\textcolor{green!50!black}{- I5: Does not introduce new \texttt{QuerySet} methods or properties beyond fixing the existing \texttt{ordered} property.}\\[0.3em]

\textcolor{green!50!black}{- R1: Ensures \texttt{ordered} returns \texttt{False} when query execution would produce SQL without \texttt{ORDER BY} clause due to \texttt{GROUP BY} with \texttt{Meta.ordering}.}\\
\textcolor{green!50!black}{- R2: Maintains \texttt{ordered} as deterministic based on query state without database round-trips or SQL compilation.}\\
\textcolor{green!50!black}{- R3: Preserves the \texttt{ordered} contract for \texttt{first} and \texttt{last} methods in \texttt{django/db/models/query.py}.}\\
\textcolor{red!70!black}{- R4: Ensures \texttt{ordered} checks \texttt{query.group\_by} state consistently whether it is \texttt{None}, \texttt{True}, or a tuple of expressions.}\\
\textcolor{red!70!black}{- R5: Avoids race conditions by checking immutable query attributes like \texttt{group\_by} and \texttt{default\_ordering} without modifying them.}\\
\textcolor{green!50!black}{- R6: Returns correct \texttt{ordered} value for chained queryset operations like \texttt{annotate} followed by \texttt{filter} without requiring SQL generation.}\\[0.3em]

\textcolor{green!50!black}{- SA1: Ensures \texttt{QuerySet.ordered} returns \texttt{False} when \texttt{annotate} with \texttt{Count} or similar aggregation creates a \texttt{GROUP BY} clause on models with \texttt{Meta.ordering}.}\\
\textcolor{red!70!black}{- SA2: Maintains \texttt{QuerySet.ordered} returning \texttt{True} when no \texttt{GROUP BY} is present even if the model has \texttt{Meta.ordering}.}\\
\textcolor{green!50!black}{- SA3: Ensures \texttt{QuerySet.ordered} still returns \texttt{True} when explicit \texttt{order\_by} is called even with \texttt{GROUP BY} present.}\\
\textcolor{red!70!black}{- SA4: Aligns \texttt{ordered} behavior with actual SQL generation where \texttt{compiler.py} lines 590--591 clear \texttt{order\_by} when grouping and \texttt{self.\_meta\_ordering} are both present.}\\
\textcolor{red!70!black}{- SA5: Preserves backward compatibility for \texttt{QuerySet.ordered} when \texttt{query.group\_by} is \texttt{None}.}\\

\vspace{0.3em}
\end{tcolorbox}

\subsection{Prompts - Baselines and Agentic Rubrics}

\label{sec:agent-prompts}

\begin{tcolorbox}[
    enhanced,
    attach boxed title to top left={xshift=6mm,yshift=-2mm},
    colback=orange-web!10,
    colframe=orange-web!50,
    colbacktitle=orange-web!70,
    title=Agentic Patch Similarity (Rollout Generation),
    fonttitle=\bfseries\color{black},
    boxed title style={size=small,colframe=orange-web,sharp corners},
    sharp corners,
    breakable
]
\begin{lstlisting}[
    basicstyle=\ttfamily\scriptsize,
    breaklines=true,
    columns=fullflexible,
    keepspaces=true,
    literate={•}{\textbullet}1
]
agent:
  templates:
    system_template: |-
      You are a helpful assistant that can interact with a computer to solve tasks.
    instance_template: |-
      <uploaded_files>
      {{working_dir}}
      </uploaded_files>
      I've uploaded a python code repository in the directory {{working_dir}}. Consider the following PR description:

      <pr_description>
      {{problem_statement}}
      </pr_description>

      Can you help me implement the necessary changes to the repository so that the requirements specified in the <pr_description> are met?
      I've already taken care of all changes to any of the test files described in the <pr_description>. This means you DON'T have to modify the testing logic or any of the tests in any way!
      Your task is to make the minimal changes to non-tests files in the {{working_dir}} directory to ensure the <pr_description> is satisfied.
      Follow these steps to resolve the issue:
      1. As a first step, it might be a good idea to find and read code relevant to the <pr_description>
      2. Create a script to reproduce the error and execute it with `python <filename.py>` using the bash tool, to confirm the error
      3. Edit the sourcecode of the repo to resolve the issue
      4. Rerun your reproduce script and confirm that the error is fixed!
      5. Think about edgecases and make sure your fix handles them as well
      Your thinking should be thorough and so it's fine if it's very long.
    next_step_template: |-
      OBSERVATION:
      {{observation}}
    next_step_no_output_template: |-
      Your command ran successfully and did not produce any output.
  tools:
    env_variables:
      PAGER: cat
      MANPAGER: cat
      LESS: -R
      PIP_PROGRESS_BAR: 'off'
      TQDM_DISABLE: '1'
      GIT_PAGER: cat
    bundles:
      - path: tools/registry
      - path: tools/edit_anthropic
      - path: tools/review_on_submit_m
    registry_variables:
      USE_FILEMAP: 'true'
      SUBMIT_REVIEW_MESSAGES:
        - |
          Thank you for your work on this issue. Please carefully follow the steps below to help review your changes.

          1. If you made any changes to your code after running the reproduction script, please run the reproduction script again.
            If the reproduction script is failing, please revisit your changes and make sure they are correct.
            If you have already removed your reproduction script, please ignore this step.
          2. Remove your reproduction script (if you haven't done so already).
          3. If you have modified any TEST files, please revert them to the state they had before you started fixing the issue.
            You can do this with `git checkout -- /path/to/test/file.py`. Use below <diff> to find the files you need to revert.
          4. Run the submit command again to confirm.

          Here is a list of all of your changes:

          <diff>
          {{diff}}
          </diff>
    enable_bash_tool: true
    disable_image_processing: true
    parse_function:
      type: function_calling
  history_processors:
    - type: cache_control
      last_n_messages: 2
  model:
    temperature: 1.
    retry:
      retries: 3
\end{lstlisting}
\end{tcolorbox}

\begin{tcolorbox}[
    enhanced,
    attach boxed title to top left={xshift=6mm,yshift=-2mm},
    colback=orange-web!10,
    colframe=orange-web!50,
    colbacktitle=orange-web!70,
    title=Agentic Patch Similarity (Judge),
    fonttitle=\bfseries\color{black},
    boxed title style={size=small,colframe=orange-web,sharp corners},
    sharp corners,
    breakable
]
\begin{lstlisting}[
    basicstyle=\ttfamily\scriptsize,
    breaklines=true,
    columns=fullflexible,
    keepspaces=true,
    literate={•}{\textbullet}1
]
REFERENCE_BASELINE_SYSTEM_PROMPT = """You are an expert code reviewer evaluating AI-generated patches for software engineering tasks. Your task is to compare a CANDIDATE PATCH against a REFERENCE PATCH (golden answer) and rate the candidate on a scale of 1-5.

Scoring Criteria:
- 5 (Excellent): The candidate patch is functionally equivalent to the reference. It correctly addresses the problem, handles edge cases, and follows best practices.
- 4 (Good): The candidate patch addresses the core problem correctly but may have minor differences in implementation approach or style compared to the reference.
- 3 (Acceptable): The candidate patch partially addresses the problem. It may miss some edge cases or have incomplete fixes, but demonstrates understanding of the issue.
- 2 (Poor): The candidate patch shows an attempt to fix the problem but has significant issues. It may introduce bugs, miss the root cause, or have incorrect logic.
- 1 (Incorrect): The candidate patch fails to address the problem, is completely wrong, or makes changes unrelated to the issue.

Return your evaluation as JSON with a score and brief reasoning.

JSON format:
{
  "score": <integer 1-5>,
  "reasoning": "<brief explanation>"
}
"""


REFERENCE_BASELINE_USER_PROMPT = """Please evaluate the CANDIDATE PATCH by comparing it to the REFERENCE PATCH (golden answer).

[PROBLEM DESCRIPTION]
{problem_statement}

[REFERENCE PATCH (Golden Answer)]
{reference_patch}

[CANDIDATE PATCH (To Evaluate)]
{candidate_patch}

Please evaluate the CANDIDATE PATCH and return the score in JSON format only.

EVALUATION:"""

\end{lstlisting}
\end{tcolorbox}

\begin{tcolorbox}[
    enhanced,
    attach boxed title to top left={xshift=6mm,yshift=-2mm},
    colback=orange-web!10,
    colframe=orange-web!50,
    colbacktitle=orange-web!70,
    title=Agentic Rubrics,
    fonttitle=\bfseries\color{black},
    boxed title style={size=small,colframe=orange-web,sharp corners},
    sharp corners,
    breakable
]
\label{appendix:rubric_gen_config}
\begin{lstlisting}[
    basicstyle=\ttfamily\scriptsize,
    breaklines=true,
    columns=fullflexible,
    keepspaces=true,
    literate={•}{\textbullet}1
]
agent:
  templates:
    system_template: |-
      You are an expert code reviewer that can understand issues and are well versed in the codebase. Your job is to write high-quality rubrics to grade the solution to a given issue.

      IMPORTANT: In EVERY turn, you MUST ALWAYS include:
      1. A summary of your thinking - explain what you're planning to do and why, and what tool you're going to use in (4-5 sentences max).
      2. A tool call. You can only make one tool call per turn.
    instance_template: |-
      <uploaded_files>
      {{working_dir}}
      </uploaded_files>
     I've uploaded a python code repository in the directory {{working_dir}}. Consider the following PR description:

      <pr_description>
      {{problem_statement}}
      </pr_description>

      Can you help me write high quality rubrics to grade the solution to the task described in the <pr_description>?
      This means you SHOULDN'T attempt to solve the task yourself, but rather understand the task, go through the codebase, and write rubrics only.
      Follow these steps to write the rubrics:
      1. As a first step, it might be a good idea to find and read code relevant to the <pr_description> by searching the codebase using search tools.
      2. Then think of the approach to solve the task (functional requirements, non-functional requirements, etc.)
      3. Also understand how the codebase is structured and how the code is organized, as well as coding style, etc.
      4. Write a list of rubrics that can be used to grade the solution to the task described in the <pr_description>. Your rubrics should be along the axes described below. 
      5. Then finally, make a new file in the {{working_dir}} directory called `rubrics.yaml` with the rubrics you wrote. It should be a valid YAML file with the structure described later below
      6. Also take a turn to ensure the yaml file is parseable by the yaml.safe_load function on the file itself. If it throws an error, fix the yaml file and take another turn to ensure it is parseable.
      7. This should be the only file you create in the {{working_dir}} directory. DO NOT create/modify/delete any other files or directories.
      8. Finally, submit the task. 

      Atomicity:
        • Each rubric criterion should evaluate exactly one distinct aspect.
        • Avoid bundling multiple criteria into a single rubric. Most stacked criteria with the word "and" can be broken up into multiple pieces. 

      Self-containment & specificity (strict):
        • Do NOT write generic items; bind each item to exact paths/symbols/tokens seen in this instance.
        • Never rely on cross-item references; each item stands alone with its own identifiers and patterns.
        • The judge will only have access to the problem and patch and the current rubric under evaluation, so make sure the rubric can be evaluated without any other information.

      Mutually Exclusive, Collectively Exhaustive (MECE):
        • The rubric set should be mutually exclusive and collectively exhaustive.
      
      Style constraints (strict):
        • YAML only-no prose outside YAML.
        • Avoid backslash-heavy patterns; if you absolutely must include one, double any backslashes so the YAML stays valid.
        • Each rubric description starts with a third-person singular verb (e.g., Identifies, Implements, Validates, Confirms, Avoids, Cleans up, Plans).
        • Make descriptions concrete using tokens from PATCH/PR_DESCRIPTION.
        • Each rubric item includes: id (short), description (verb-first, instance-grounded), weight (int; 1=nice, 2=valuable, 3=must).
        • Avoid double-counting: do not re-score the same behavior under multiple items.

      Pattern-writing guidelines (keep literal YAML-friendly text; no regex required):
        • Use plain path mentions like "diff --git a/path/to/file.py" or "+++ b/path/to/file.py".
        • Refer to symbols with straightforward phrases such as "def my_function(" instead of regex classes.
        • Describe value patterns in words (e.g., "string containing total") instead of complex expressions.
        • If you need to forbid something, just mention the exact string '@pytest.mark.skip', etc

      Axes (execution-free):
        • file_change_rubrics (4 - 8): Scope, locality, and sufficiency of edits in PATCH (files/symbols/guards/regexes/flags). Penalize unrelated file churn; reward minimal, reversible diffs tied to the stated bug.
        • spec_alignment_rubrics (3 - 6): Alignment of code to PR_DESCRIPTION. Use textual acceptance criteria (required types/conditions/error handling/API contracts) and ensure the patch reflects them.
        • integrity_rubrics (3 - 6): Hygiene and "no-cheating" safeguards-avoid test weakening (if tests appear in PATCH), mass renames, or dependency churn; preserve public API/semantics unless PR_DESCRIPTION requires otherwise.
        • runtime_rubrics (3 - 6): Natural-language criteria describing **intended runtime behavior** (NOT concrete tests), supported by execution-free textual evidence.
            - **Distinguishability:** Ensures the patch introduces or preserves signals that differentiate correct vs. incorrect behavior under plausible inputs (e.g., specific exception class, sentinel return, boundary guard).  
            - **Regression safety:** Confirms backward-compatibility constraints (e.g., original API signatures/flags remain valid, deprecations gated via warnings).  
            - **Determinism / flake resistance:** Avoids nondeterministic sources at runtime (unseeded randomness, wall-clock sleeps, network I/O) that would make tests flaky.  
            - **Resource & timeout bounds:** Prevents pathological loops or heavy calls; respects existing timeouts/limits.  
            - **Error-surface clarity:** Produces stable, specific messages/exception types that a test could assert against (not vague strings).  
            - **Harness integrity:** Does not bypass or disable the project's runner/verifier hooks (e.g., keeps regression filters, CLI exit codes).  

      Return exactly this YAML structure:
      metadata:
        task_summary: "<one-sentence summary grounded in PR_DESCRIPTION>"
        underlying_bug: "<precise failure trigger grounded in PR_DESCRIPTION>"
      axes:
        file_change_rubrics:
          - id: "FC1"
            description: "Identifies ..."
            weight: 3
          - id: "FC2"
            description: "Identifies ..."
            weight: 2
        spec_alignment_rubrics:
          - id: "SA1"
            description: "Recognizes ..."
            weight: 2
        integrity_rubrics:
          - id: "I1"
            description: "Confirms ..."
            weight: 2
        runtime_rubrics:
          - id: "R1"
            description: "Maintains ..."
            weight: 2
              
    next_step_template: |-
      OBSERVATION:
      {{observation}}
    next_step_no_output_template: |-
      Your command ran successfully and did not produce any output.
  tools:
    env_variables:
      PAGER: cat
      MANPAGER: cat
      LESS: -R
      PIP_PROGRESS_BAR: 'off'
      TQDM_DISABLE: '1'
      GIT_PAGER: cat
    bundles:
      - path: tools/registry
      - path: tools/edit_anthropic
      - path: tools/review_on_submit_m
    registry_variables:
      USE_FILEMAP: 'true'
      SUBMIT_REVIEW_MESSAGES:
        - |
          Thank you for your work on writing the rubrics. Please carefully follow the steps below to help review your changes.

          1. If you made any changes to your code other than the `rubrics.yaml` file in the testbed directory, please revert them to the state they had before you started writing the rubrics.
          2. You can do this with `git checkout -- /path/to/file.py`. Use below <diff> to find the files you need to revert. This has to be done, otherwise we can't extract the rubrics.yaml file.          3. Run the submit command again to confirm.

          Here is a list of all of your changes:

          <diff>
          {{diff}}
          </diff>
    enable_bash_tool: true
    disable_image_processing: true
    parse_function:
      type: function_calling
  history_processors:
    - type: cache_control
      last_n_messages: 2
  model:
    temperature: 1.0
    retry:
      retries: 3
\end{lstlisting}
\end{tcolorbox}

\begin{tcolorbox}[
    enhanced,
    attach boxed title to top left={xshift=6mm,yshift=-2mm},
    colback=orange-web!10,
    colframe=orange-web!50,
    colbacktitle=orange-web!70,
    title=Agentic Tests,
    fonttitle=\bfseries\color{black},
    boxed title style={size=small,colframe=orange-web,sharp corners},
    sharp corners,
    breakable
]
\begin{lstlisting}[
    basicstyle=\ttfamily\scriptsize,
    breaklines=true,
    columns=fullflexible,
    keepspaces=true,
    literate={•}{\textbullet}1
]
agent:
  templates:
    system_template: |-
      You are a programming agent who is provided a github issue and repository bash environment and is tasked to generate a standalone test script that can reproduce and verify the issue without relying on any testing frameworks.
    instance_template: |-
      <uploaded_files>
      {{working_dir}}
      </uploaded_files>
      I've uploaded a python code repository in the directory {{working_dir}}. Consider the following PR description:

      <pr_description>
      {{problem_statement}}
      </pr_description>

      Can you help me write a standalone test_issue.py file that tests and reproduces the issue described in the <pr_description>?
      This test file should be completely self-contained and executable directly with Python, without requiring any testing frameworks like pytest or unittest.

      IMPORTANT GUIDELINES:
      1. First, explore the repository to understand what the issue is about and how to test and reproduce it. Focus on understanding the core functionality rather than the testing structure.

      2. Create a standalone Python script (test_issue.py) that:
        - Imports only the necessary modules from the repository
        - Sets up the minimum environment needed to reproduce the issue
        - Contains all logic within the script itself (no external test dependencies)
        - Runs quickly and terminates itself (no background servers or long-running processes)
        - Write at least ten test cases to test the issue.

      3. CRITICAL: For each of the test cases: your test script MUST use these EXACT print statements to indicate test results for each test case:
        - Print "FAILED" when the code confirms the bug exists, and so the test case fails.
        - Print "PASSED" when the code runs without the issue and so the test case passes.
        - Print "Other issues" when unexpected problems occur
        IMPORTANT: Again include the above print statements for each of the test cases in /testbed/test_issue.py. 

      4. Include error handling to prevent the script from crashing:
        - Catch exceptions appropriately
        - Always output one of the three exact phrases above
        - DO NOT use assertions that might terminate the program (without error handling)

      5. The test should fail (print "FAILED") when run against the current repo state.

      6. Your test script should also check for the correct behaviour when the issue is fixed (i.e. print "PASSED"). If the issue is not fixed and the code exhibits incorrect behavior after applying a fix, it should print "Other issues" or "FAILED" as appropriate.
    
      7. Write the final test script to /testbed/test_issue.py. Ensure that the script is runnable via `python test_issue.py`.

      8. The final line in the test script should be a comment about how many test cases were written. Example: `# Total tests: 6`. 
      
      9. This is important each test case number (Out of the total you wrote), we will parse the output of test script one-by-one and check if the solution passes test case i. For example, """test_case_{i} PASSED""" or """test_case_{i} FAILED""".

      Example format for a single test case in the test script:
      ```python
      import sys
      from some_package import relevant_module

      def test1():
          try:
              # Setup minimal test environment
              test_input = "example input that triggers the issue"

              # Attempt the operation that should reproduce the issue
              result = relevant_module.function_with_issue(test_input)

              # check if the issue is reproduced
              if result == "expected output that indicates the issue":
                  return "FAILED"
              else:
                  # check if result matches the expected output when the issue is resolved
                  # ensure to perform all necessary checks
                  assert result == "expected output when resolved"
                  return "PASSED"
                  
          except Exception as e:
              return "Other issues"  # Optional: can include error details for debugging

      ...

      if __name__ == "__main__":
          print(f"test_case_1 {test1()}")
          ...

      # Total tests: 5
      ```

      FINAL CHECKS:
      - Does each one of your test run standalone (without pytest/unittest)?
      - Does each one of your test contain EXACTLY ONE of the three required print statements?
      - Does each one of your test terminate automatically after printing the result?
      - Does each one of your test properly reproduce the issue described in the problem statement?
      - Is it simple, focused, and free of unnecessary complexity?
      - Does the final line in the test script contain the correct number of test cases and with the exact format `# Total tests: <number of test cases>` (no commas, no spaces, no other text)?

      GENERAL INSTRUCTIONS:
      - Each response must include both 
        - natural language reasoning about your approach
        - a function call to solve the task
      - You can take multiple turns to solve the task, but only finish once you're confident in your solution
      - If a file_editor edit fails, view the file before retrying with adjusted content

      General Steps:
      1. Understand the issue, corresponding code and how to reproduce the issue.
      2. Write a standalone test script that reproduces the issue. Make sure that the output is "FAILED" for each of the single test.
      3. Add further test cases including more thorough testing, inputs, edge cases to ensure the issue is correctly identified.
      4. Run the test script to ensure output is as expected (see example output format below).

      The final output of the test script should resemble the following format (just an example):
      <EXAMPLE OUTPUT FORMAT>
      test_case_1 FAILED
      test_case_2 PASSED 
      test_case_3 FAILED
      test_case_4 PASSED
      test_case_5 FAILED
      test_case_6 PASSED
      test_case_7 FAILED
      test_case_8 PASSED
      test_case_9 FAILED
      </EXAMPLE OUTPUT FORMAT>
      You must follow the above format for the output of the test script. Other issues should be max 1-2 test cases (in worst case).

      Finally, use submit tool to submit.
      
      CRITICAL: Do not submit until you have added diverse test cases and thoroughly verified the output of the test script.
      NOTE: for django environments: you should use test_sqlite settings file during testing.

    next_step_template: |-
      OBSERVATION:
      {{observation}}
    next_step_no_output_template: |-
      Your command ran successfully and did not produce any output.
  tools:
    env_variables:
      PAGER: cat
      MANPAGER: cat
      LESS: -R
      PIP_PROGRESS_BAR: 'off'
      TQDM_DISABLE: '1'
      GIT_PAGER: cat
    bundles:
      - path: tools/registry
      - path: tools/edit_anthropic
      - path: tools/review_on_submit_m
    registry_variables:
      USE_FILEMAP: 'true'
      SUBMIT_REVIEW_MESSAGES:
        - |
          Thank you for your work on writing the tests. Please carefully follow the steps below to help review your changes.

          1. If you made any changes to your code other than the `test_issue.py` file, please revert them to the state they had before you started writing the tests.
          2. You can do this with `git checkout -- /path/to/file.py`. Use below <diff> to find the files you need to revert.
          3. Run the submit command again to confirm.

          Here is a list of all of your changes:

          <diff>
          {{diff}}
          </diff>
    enable_bash_tool: true
    disable_image_processing: true
    parse_function:
      type: function_calling
  history_processors:
    - type: cache_control
      last_n_messages: 2
  model:
    temperature: 1.
    retry:
      retries: 3

\end{lstlisting}
\end{tcolorbox}

\begin{tcolorbox}[
    enhanced,
    attach boxed title to top left={xshift=6mm,yshift=-2mm},
    colback=orange-web!10,
    colframe=orange-web!50,
    colbacktitle=orange-web!70,
    title=Patch Classifier,
    fonttitle=\bfseries\color{black},
    boxed title style={size=small,colframe=orange-web,sharp corners},
    sharp corners,
    breakable
]
\begin{lstlisting}[
    basicstyle=\ttfamily\scriptsize,
    breaklines=true,
    columns=fullflexible,
    keepspaces=true,
    literate={•}{\textbullet}1
]
SYSTEM_PROMPT = """You are an expert judge evaluating AI assistant interactions. Your task is to determine if the assistant successfully resolved the user's request.

Key evaluation criteria:
1. Did the assistant complete the main task requested by the user?
2. Did the assistant handle all edge cases and requirements specified?
3. Were there any errors or issues in the final solution?

Respond only with "<judgement>YES</judgement>" or "<judgement>NO</judgement>" based on
if the assistant successfully resolved the user's request."""

USER_PROMPT_TEMPLATE = """Please evaluate the following request to solve a coding issue and the proposed solution:

[PROMPT]

{problem_statement}

[SOLUTION]

{model_patch}"""

\end{lstlisting}
\end{tcolorbox}

\begin{tcolorbox}[
    enhanced,
    attach boxed title to top left={xshift=6mm,yshift=-2mm},
    colback=orange-web!10,
    colframe=orange-web!50,
    colbacktitle=orange-web!70,
    title=Non-Agentic Rubrics,
    fonttitle=\bfseries\color{black},
    boxed title style={size=small,colframe=orange-web,sharp corners},
    sharp corners,
    breakable
]
\begin{lstlisting}[
    basicstyle=\ttfamily\scriptsize,
    breaklines=true,
    columns=fullflexible,
    keepspaces=true,
    literate={•}{\textbullet}1
]
SYSTEM_PROMPT = """You are an expert code reviewer that can understand issues and are well versed in codebases. Your job is to write high-quality rubrics to grade the solution to a given issue.

Based on the problem description provided, write rubrics that can be used to evaluate a patch that attempts to solve the issue.

Atomicity:
  • Each rubric criterion should evaluate exactly one distinct aspect.
  • Avoid bundling multiple criteria into a single rubric. Most stacked criteria with the word "and" can be broken up into multiple pieces. 

Self-containment & specificity (strict):
  • Do NOT write generic items; bind each item to exact paths/symbols/tokens mentioned in the problem description.
  • Never rely on cross-item references; each item stands alone with its own identifiers and patterns.
  • The judge will only have access to the problem and patch and the current rubric under evaluation, so make sure the rubric can be evaluated without any other information.

Mutually Exclusive, Collectively Exhaustive (MECE):
  • The rubric set should be mutually exclusive and collectively exhaustive.

Style constraints (strict):
  • YAML only-no prose outside YAML.
  • Avoid backslash-heavy patterns; if you absolutely must include one, double any backslashes so the YAML stays valid.
  • Each rubric description starts with a third-person singular verb (e.g., Identifies, Implements, Validates, Confirms, Avoids, Cleans up, Plans).
  • Make descriptions concrete using tokens from the problem description.
  • Each rubric item includes: id (short), description (verb-first, instance-grounded), weight (int; 1=nice, 2=valuable, 3=must).
  • Avoid double-counting: do not re-score the same behavior under multiple items.

Pattern-writing guidelines (keep literal YAML-friendly text; no regex required):
  • Use plain path mentions like "diff --git a/path/to/file.py" or "+++ b/path/to/file.py".
  • Refer to symbols with straightforward phrases such as "def my_function(" instead of regex classes.
  • Describe value patterns in words (e.g., "string containing total") instead of complex expressions.
  • If you need to forbid something, just mention the exact string '@pytest.mark.skip', etc

Axes (execution-free):
  • file_change_rubrics (4-8): Scope, locality, and sufficiency of edits in PATCH (files/symbols/guards/regexes/flags). Penalize unrelated file churn; reward minimal, reversible diffs tied to the stated bug.
  • spec_alignment_rubrics (3-6): Alignment of code to the problem description. Use textual acceptance criteria (required types/conditions/error handling/API contracts) and ensure the patch reflects them.
  • integrity_rubrics (3-6): Hygiene and "no-cheating" safeguards-avoid test weakening (if tests appear in PATCH), mass renames, or dependency churn; preserve public API/semantics unless the problem description requires otherwise.
  • runtime_rubrics (3-6): Natural-language criteria describing **intended runtime behavior** (NOT concrete tests), supported by execution-free textual evidence.
      - **Distinguishability:** Ensures the patch introduces or preserves signals that differentiate correct vs. incorrect behavior under plausible inputs (e.g., specific exception class, sentinel return, boundary guard).  
      - **Regression safety:** Confirms backward-compatibility constraints (e.g., original API signatures/flags remain valid, deprecations gated via warnings).  
      - **Determinism / flake resistance:** Avoids nondeterministic sources at runtime (unseeded randomness, wall-clock sleeps, network I/O) that would make tests flaky.  
      - **Resource & timeout bounds:** Prevents pathological loops or heavy calls; respects existing timeouts/limits.  
      - **Error-surface clarity:** Produces stable, specific messages/exception types that a test could assert against (not vague strings).  
      - **Harness integrity:** Does not bypass or disable the project's runner/verifier hooks (e.g., keeps regression filters, CLI exit codes).  

Return exactly this YAML structure (and nothing else):
metadata:
  task_summary: "<one-sentence summary grounded in the problem description>"
  underlying_bug: "<precise failure trigger grounded in the problem description>"
axes:
  file_change_rubrics:
    - id: "FC1"
      description: "Identifies ..."
      weight: 3
    - id: "FC2"
      description: "Identifies ..."
      weight: 2
  spec_alignment_rubrics:
    - id: "SA1"
      description: "Recognizes ..."
      weight: 2
  integrity_rubrics:
    - id: "I1"
      description: "Confirms ..."
      weight: 2
  runtime_rubrics:
    - id: "R1"
      description: "Maintains ..."
      weight: 2
"""

USER_PROMPT_TEMPLATE = """Consider the following problem description:

<problem_description>
{problem_statement}
</problem_description>

Write high quality rubrics to grade a patch that attempts to solve the task described in the <problem_description>.
Output ONLY valid YAML with the structure specified in the system prompt. Do not include any other text or explanation.
"""
\end{lstlisting}
\end{tcolorbox}

\subsection{Rubric Judge Prompt}

\begin{tcolorbox}[
    enhanced,
    attach boxed title to top left={xshift=6mm,yshift=-2mm},
    colback=sagegreen!30,
    colframe=sagegreen!50,
    colbacktitle=sagegreen!70,
    title=Rubric Judge Prompt,
    fonttitle=\bfseries\color{black},
    boxed title style={size=small,colframe=sagegreen,sharp corners},
    sharp corners,
    breakable
]
\begin{lstlisting}[
    basicstyle=\ttfamily\scriptsize,
    breaklines=true,
    columns=fullflexible,
    keepspaces=true,
    literate={•}{\textbullet}1
]
SYSTEM_PROMPT = """
You are a rubric based evaluator for software-engineering agent's generated patch. Use the provided rubric to evaluate the generated patch.

Inputs (provided later):
	- PR_DESCRIPTION: problem + expected behavior.
	- RUBRIC: dictionary of n rubric items for an ideal patch with their ids as keys and descriptions as values.
  - PATCH: the model's predicted code patch 

Your job:
	1.	Analyze the rubric and the patch to evaluate the SWE-agent's generated patch.
	2.	Emit a score for each rubric item. The score should be a binary score of 1 if the patch satisfies the rubric item and 0 otherwise.

Return the scores in a JSON format.

JSON format:
{
  "<rubric_id_1>": <score_1>,
  "<rubric_id_2>": <score_2>,
  "<rubric_id_3>": <score_3>,
  "<rubric_id_4>": <score_4>,
  ...
  "<rubric_id_n>": <score_n>
}"""


USER_PROMPT = """
PR_DESCRIPTION:
{pr_description}

PATCH:
{patch}

RUBRIC:
{rubric}

Please evaluate the PATCH using the rubric and return the scores in JSON format.
SCORES:
"""
\end{lstlisting}
\end{tcolorbox}

\subsection{Rubric Utility Analysis Prompt}
\label{sec:utility-prompts}
\begin{tcolorbox}[
    enhanced,
    attach boxed title to top left={xshift=6mm,yshift=-2mm},
    colback=sagegreen!30,
    colframe=sagegreen!50,
    colbacktitle=sagegreen!70,
    title=Rubric Utility Analysis Prompt,
    fonttitle=\bfseries\color{black},
    boxed title style={size=small,colframe=sagegreen,sharp corners},
    sharp corners,
    breakable
]
\begin{lstlisting}[
    basicstyle=\ttfamily\scriptsize,
    breaklines=true,
    columns=fullflexible,
    keepspaces=true
]
SYSTEM_PROMPT = """
You are an expert software engineer. Your job is to analyze how a candidate patch
is graded by tests and rubrics.

Inputs (provided later):
  - problem_statement
  - golden_patch (the ground-truth code patch)
  - candidate_patch (the model's predicted code patch)
  - golden_test_cases
  - test_case_reward (0 or 1)
  - rubric_descriptions (map: rubric_id -> text)
  - rubrics_to_analyze (list of rubric ids to analyze)

Treat problem_statement + golden_patch + golden_test_cases as the ground-truth
specification for correct behavior.

We focus on HIGH-ALIGNMENT cases between tests and rubrics:
  - If test_case_reward = 0, rubrics_to_analyze is the set of FAILING rubrics.
    Here we ask: how well do these failing rubrics align with the reasons the
    golden tests reject the candidate_patch?
  - If test_case_reward = 1, rubrics_to_analyze is the set of ACCEPTED rubrics.
    Here we ask: how well do these accepted rubrics align with the reasons the
    golden tests accept the candidate_patch?

Your task:
  1. For each rubric in rubrics_to_analyze, decide whether it is:
     a) Valid   - its judgment is consistent with the ground-truth spec and it
                  provides a correct, meaningful reason that agrees with the
                  test outcome.
     b) Spurious - its judgment is not well supported by the ground-truth spec
                  or adds noise (e.g., unnecessary constraints, conflicts with
                  the golden behavior, or misinterprets the situation), even
                  though the sign of the score aligns with the tests.
  2. Assign each rubric to exactly ONE sub-category under its main category
     (Valid or Spurious). If no sub-category fits, create a new one with a short
     title and a brief description.

VALID sub-categories (use when the rubric adds real value and is consistent
with the spec):
  - Core Semantics
      Checks whether the patch actually fixes or preserves the functional
      behavior described in the problem (root cause, outputs, semantics).
  - Edge Coverage
      Enforces handling of important edge cases or reproducer scenarios that
      are implied by the spec but not fully covered by tests.
  - API / Compat
      Ensures public API shape, types, and behavior stay compatible with
      existing callers/versions, or change exactly as required by the spec.
  - Structure / Scope
      Ensures the change is in the correct module/layer and reasonably
      localized (no wrong-layer fix, no unrelated edits, no scope creep).
  - Performance Risk
      Flags likely performance or resource regressions that tests do not
      directly measure.
  - Security / Safety
      Enforces validation, security, or safety properties that should not be
      weakened by the patch.
  - [NEW VALID]
      If none fit, create a new Valid sub-category with a short title and a
      1-2 sentence description.

SPURIOUS sub-categories (use when the patch is acceptable given the spec, but
the rubric's failure is not well justified):
  - Low-Signal
      Encodes irrelevant or redundant constraints (style-only, hygiene-only,
      or no new information beyond other checks).
  - Over-Specified
      Demands a specific implementation even though alternative correct fixes
      are allowed.
  - Spec Conflict
      Contradicts the problem statement, golden_patch, or golden_test_cases
      (forbids behavior the spec allows or disagrees with the reference).
  - Test Mismatch
      Assumes a different test setup or evaluation protocol than the one
      actually used (e.g., expects test edits that are not in scope).
  - Eval Error
      Failure is caused by a scoring / matching / parsing bug rather than a
      real property of the candidate_patch.
  - [NEW SPURIOUS]
      If none fit, create a new Spurious sub-category with a short title and a
      1-2 sentence description.

Reasoning rules:
  - Always treat golden_patch + golden_test_cases as the authoritative spec.
  - If a rubric's explanation clearly conflicts with this spec, it is almost
    certainly Spurious.
  - Mark a rubric as Valid only if its judgment and rationale are consistent
    with the spec AND meaningfully explain why the test outcome (0 or 1) is
    correct.

Output format:
Return a JSON array, one entry per rubric in rubrics_to_analyze:

[
  {
    "rubric_id": "FC5",
    "rubric_description": "<from rubric_descriptions>",
    "tier_category": "Valid" or "Spurious",
    "subcategory_title": "<one of the titles above or a new one>",
    "subcategory_description": "<1-2 sentence definition>",
    "justification": "<2-4 sentences citing candidate_patch, golden_patch,
                      golden_test_cases, and referring to test_case_reward>"
  }
]
"""

USER_PROMPT = """
problem_statement:
{problem_statement}

golden_patch:
{golden_patch}

candidate_patch:
{candidate_patch}

golden_test_cases:
{golden_test_cases}

test_case_reward:
{test_case_reward}

rubric_descriptions:
{rubric_descriptions}

rubrics_to_analyze:
{rubrics_to_analyze}

Please classify each rubric in rubrics_to_analyze as Valid or Spurious,
assign a sub-category, and return the JSON array as specified above.

RESULT:
"""
\end{lstlisting}
\end{tcolorbox}

\end{document}